\documentclass[final]{cvpr}

\usepackage{times}
\usepackage{epsfig}
\usepackage{graphicx}
\usepackage{amsmath}
\usepackage{amssymb}
\usepackage{multirow}
\usepackage[pagebackref=true,breaklinks=true,colorlinks,bookmarks=false]{hyperref}
\def\cvprPaperID{1728} % *** Enter the CVPR Paper ID here
\def\confYear{CVPR 2021}

\begin{document}

%%%%%%%%% TITLE
\title{CFNet: Cascade and Fused Cost Volume for Robust Stereo Matching}

\author{Zhelun Shen\textsuperscript{1,2}, Yuchao Dai\textsuperscript{1}\footnotemark[1],  Zhibo Rao\textsuperscript{1}\\
\textsuperscript{1}Northwestern Polytechnical University, Xi'an, China \textsuperscript{2}Peking University, Beijing, China\\
{\tt\small shenzhelun@pku.edu.cn, daiyuchao@nwpu.edu.cn, raoxi36@foxmail.com}
%\and
%Yuchao Dai\\ \\
%{\tt\small daiyuchao@nwpu.edu.cn}
%\and
%Zhibo Rao\\ \\
%{\tt\small raoxi36@foxmail.com}
}

\maketitle

\pagestyle{empty}
\thispagestyle{empty}

\renewcommand{\thefootnote}{\fnsymbol{footnote}}

\footnotetext[1]{Yuchao Dai is the corresponding author.}
%%%%%%%%% ABSTRACT
\begin{abstract}
Recently, the ever-increasing capacity of large-scale annotated datasets has led to profound progress in stereo matching. However, most of these successes are limited to a specific dataset and cannot generalize well to other datasets. The main difficulties lie in the large domain differences and unbalanced disparity distribution across a variety of datasets, which greatly limit the real-world applicability of current deep stereo matching models. In this paper, we propose CFNet, a Cascade and Fused cost volume based network to improve the robustness of the stereo matching network. First, we propose a fused cost volume representation to deal with the large domain difference. By fusing multiple low-resolution dense cost volumes to enlarge the receptive field, we can extract robust structural representations for initial disparity estimation. Second, we propose a cascade cost volume representation to alleviate the unbalanced disparity distribution. Specifically, we employ a variance-based uncertainty estimation to adaptively adjust the next stage disparity search space, in this way driving the network progressively prune out the space of unlikely correspondences. By iteratively narrowing down the disparity search space and improving the cost volume resolution, the disparity estimation is gradually refined in a coarse-to-fine manner. When trained on the same training images and evaluated on KITTI, ETH3D, and Middlebury datasets with the fixed model parameters and hyperparameters, our proposed method achieves the state-of-the-art overall performance and obtains the 1st place on the stereo task of Robust Vision Challenge 2020.
The code will be available at \url{https://github.com/gallenszl/CFNet}.

%\YD{A key problem in this paper: how does the proposed algorithm achieve robust stereo matching.}
%These approaches generally show their superiority on a specific dataset by directly comparing the results of dozens of methods on that dataset. 

\end{abstract}

%%%%%%%%% BODY TEXT
\section{Introduction}

\begin{figure}[!htb]
 \centering
 \tabcolsep=0.05cm
 \begin{tabular}{c c c}
    \includegraphics[width=0.33\linewidth]{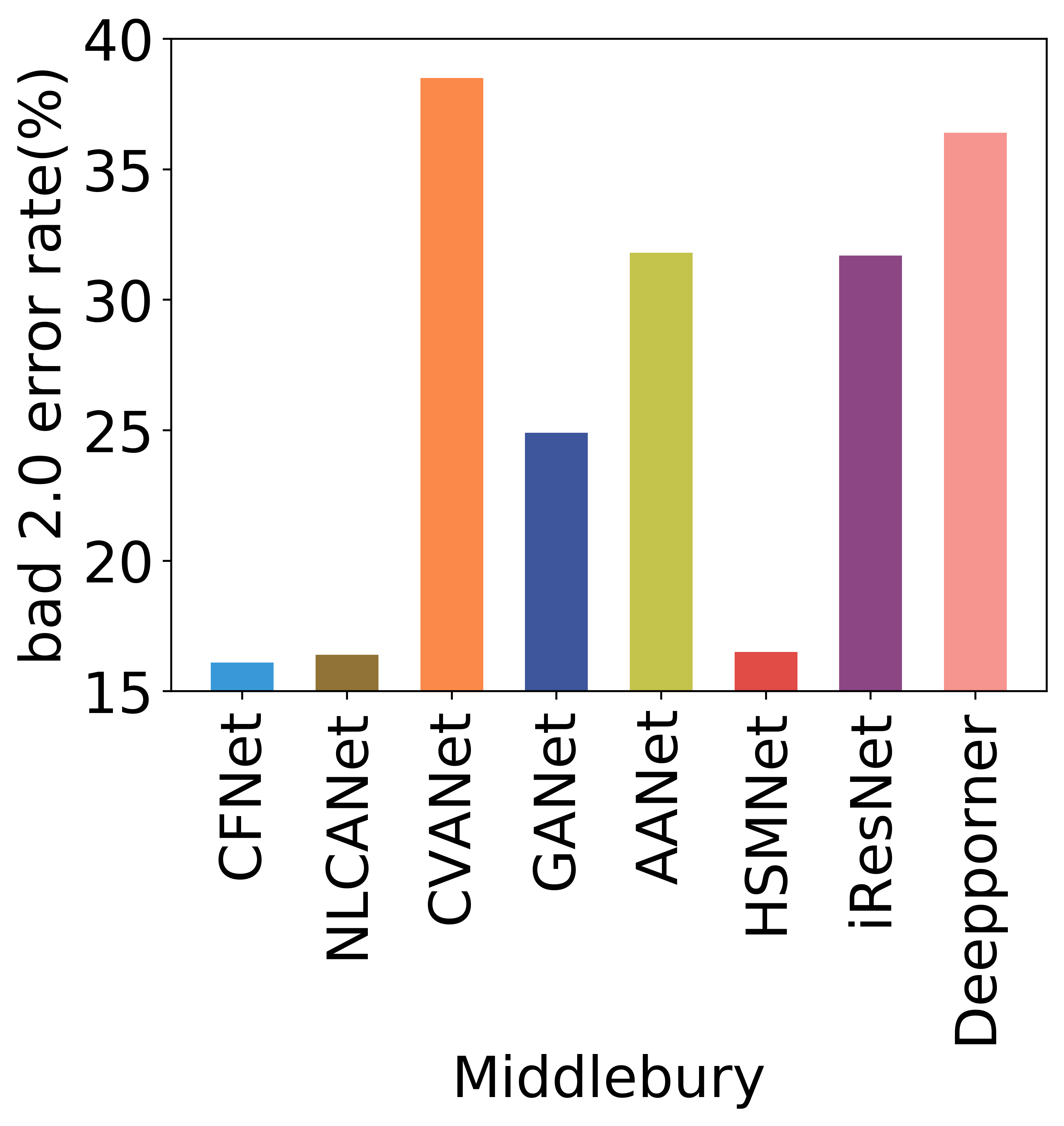}&
	\includegraphics[width=0.33\linewidth]{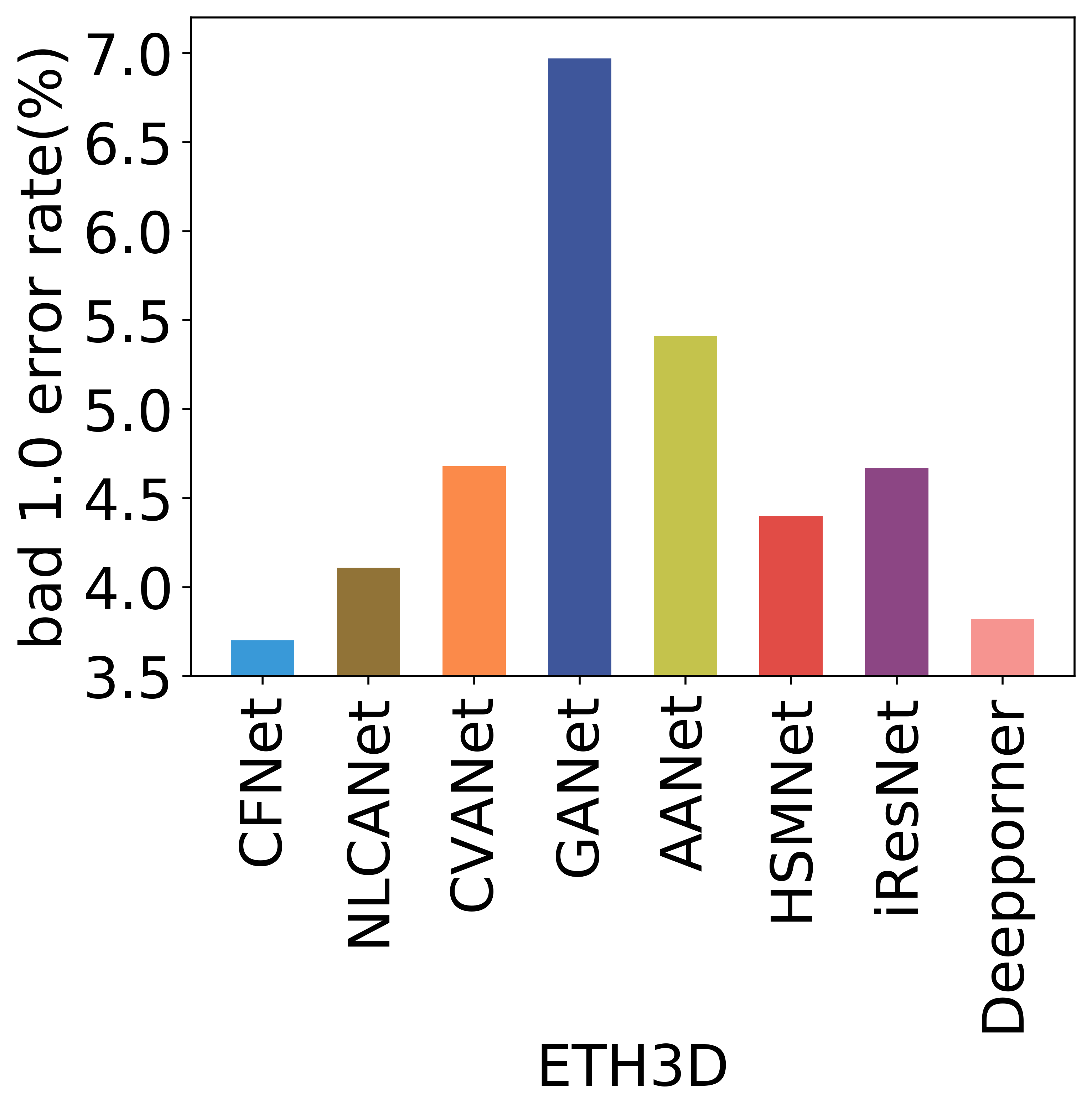}&
	\includegraphics[width=0.33\linewidth]{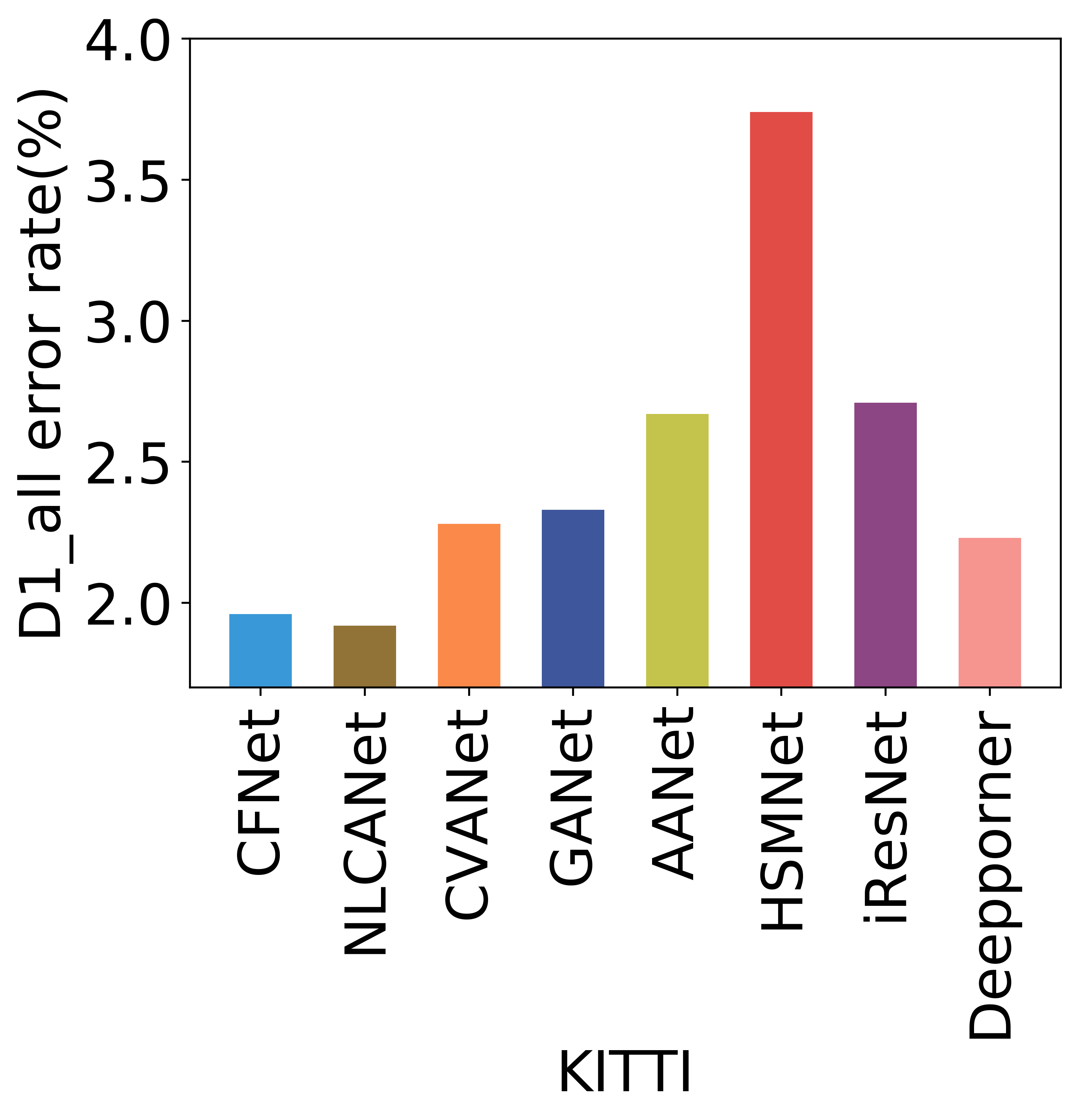}\\
	\end{tabular}
 \caption{Performance comparison in terms of generalization ability on the Middlebury, ETH3D, and KITTI 2015 datasets. Bad 2.0, bad1.0, and D1\_all (the lower the better) are used for evaluation. All methods are trained on the same training images and tested on three datasets with single model parameters and hyper-parameters. Our CFNet achieves state-of-the-art generalization and performs well on all three real-world datasets.}
 \label{fig: robust comparison}
\end{figure}

Stereo matching, \ie estimating a disparity/depth map from a pair of stereo images, is fundamental to various applications such as autonomous driving \cite{autonomousdriving}, robot navigation \cite{roboticsnavigation}, SLAM \cite{slam1,slam2}, \etc. Recently, many deep learning-based stereo methods have been developed and achieved impressive performance on most of the standard benchmarks.

\begin{figure*}[!htb]
	\centering
	\tabcolsep=0.05cm
	\begin{tabular}{c c c c}
    \includegraphics[width=0.21\linewidth]{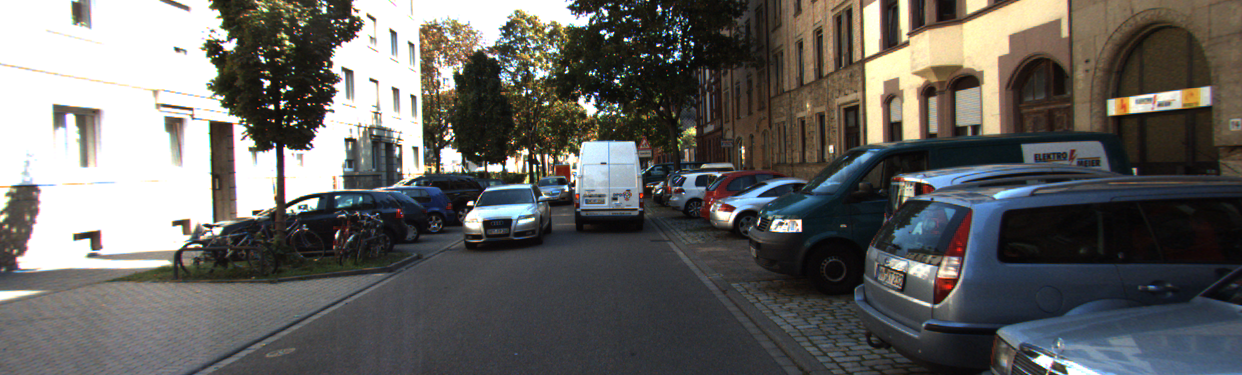}&
	\includegraphics[width=0.21\linewidth]{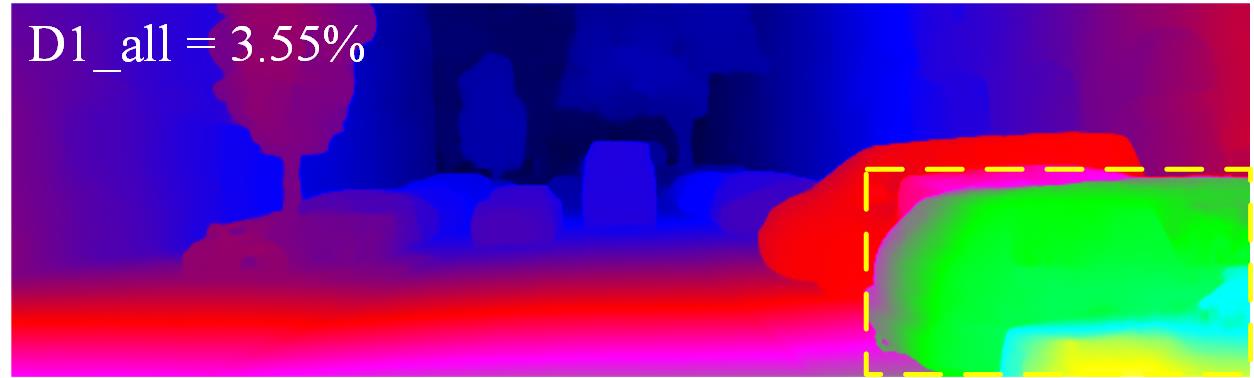}&
	\includegraphics[width=0.21\linewidth]{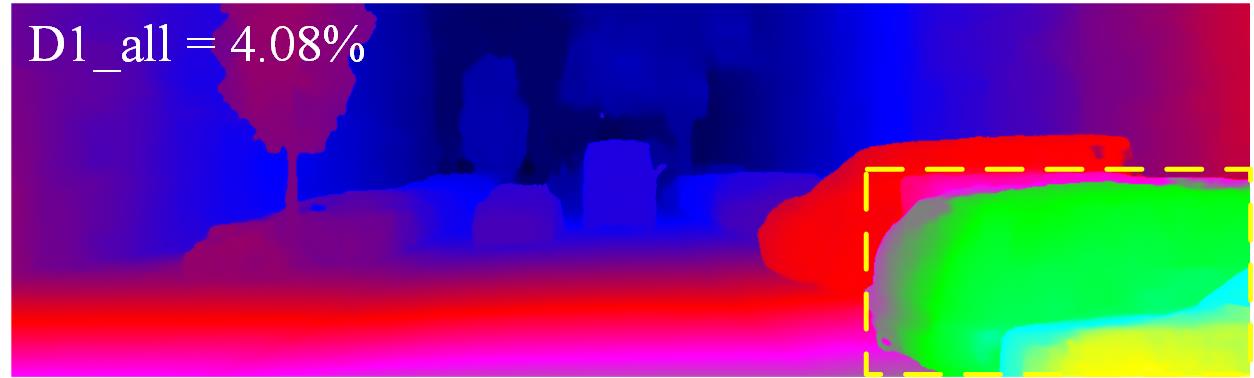}&
	\includegraphics[width=0.21\linewidth]{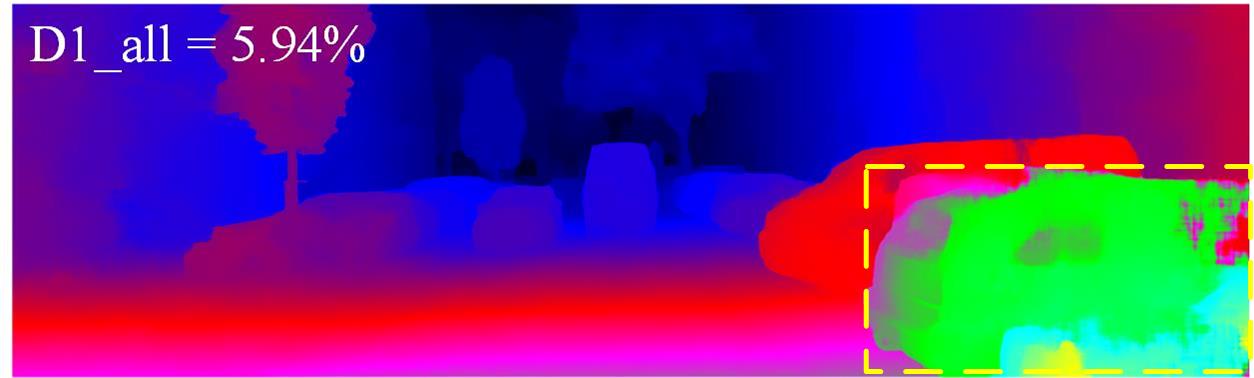}\\
	
    \includegraphics[width=0.21\linewidth]{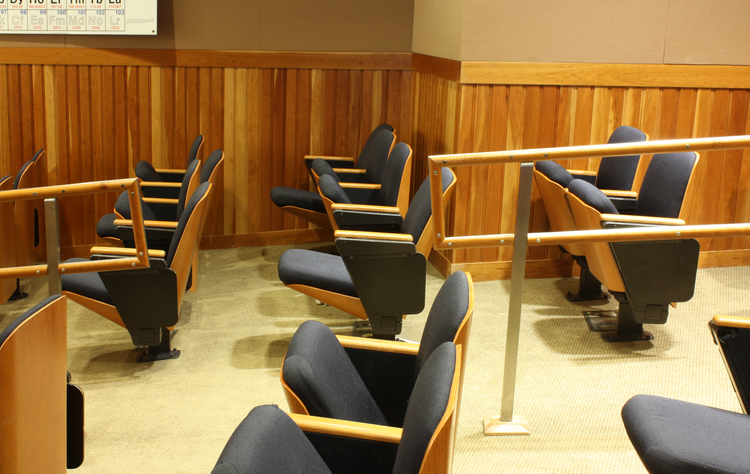}&
    \includegraphics[width=0.21\linewidth]{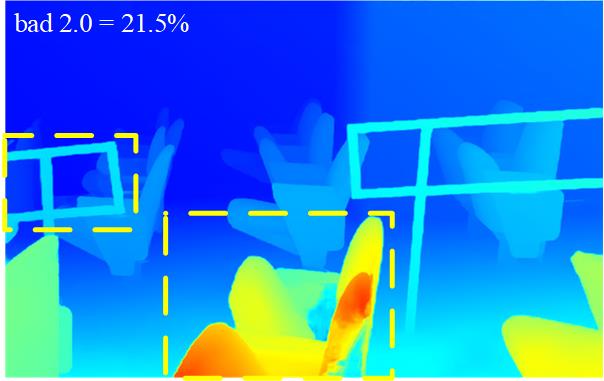}&
	\includegraphics[width=0.21\linewidth]{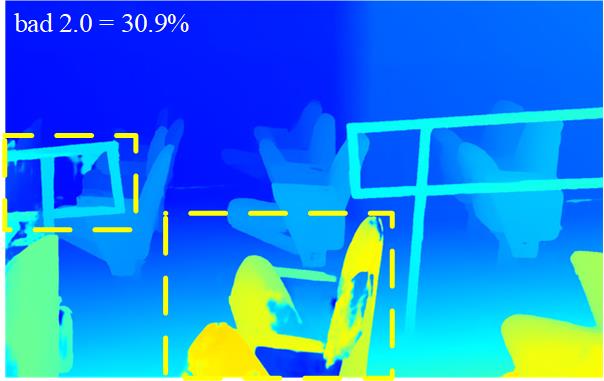}&
	\includegraphics[width=0.21\linewidth]{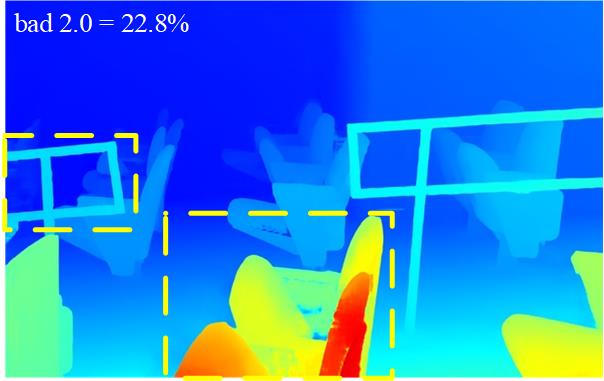}\\

    \includegraphics[width=0.21\linewidth]{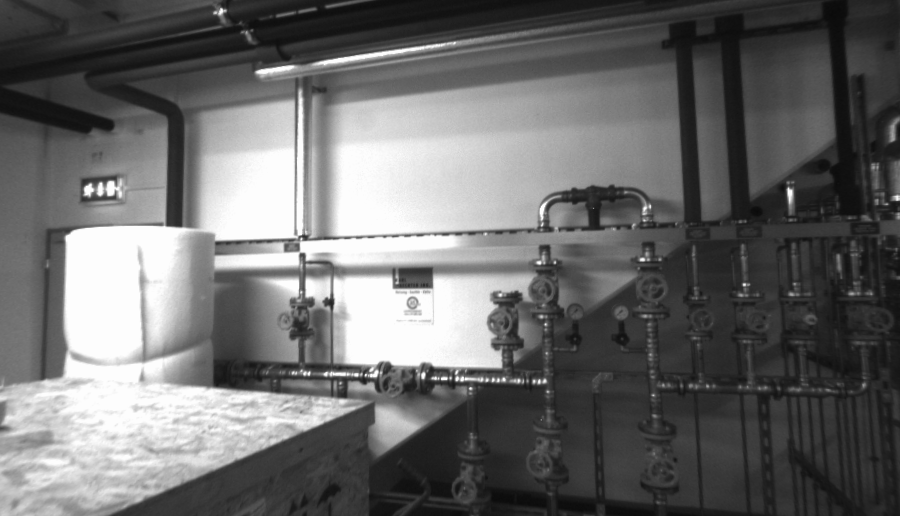}&
    \includegraphics[width=0.21\linewidth]{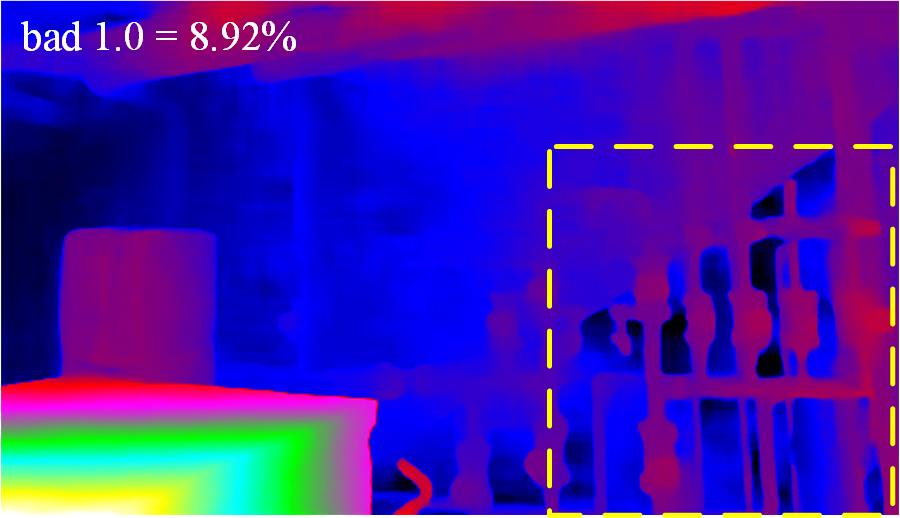}&
	\includegraphics[width=0.21\linewidth]{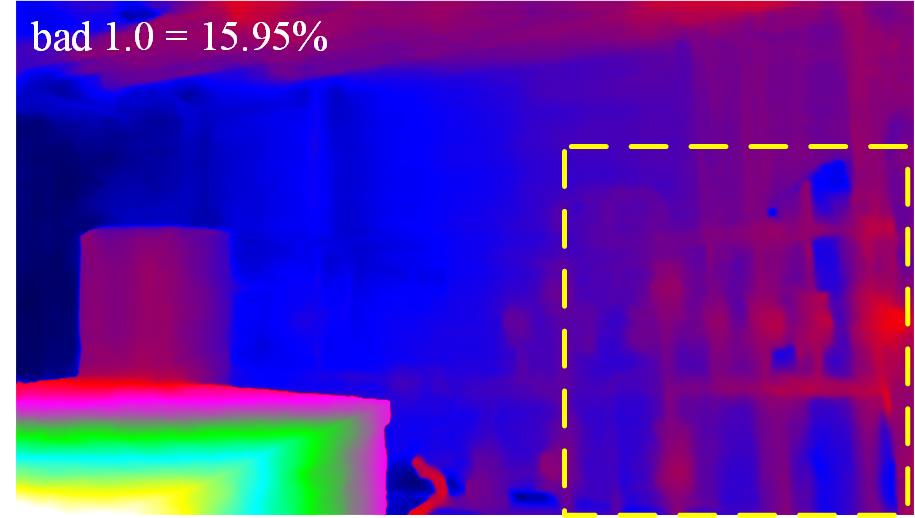}&
	\includegraphics[width=0.21\linewidth]{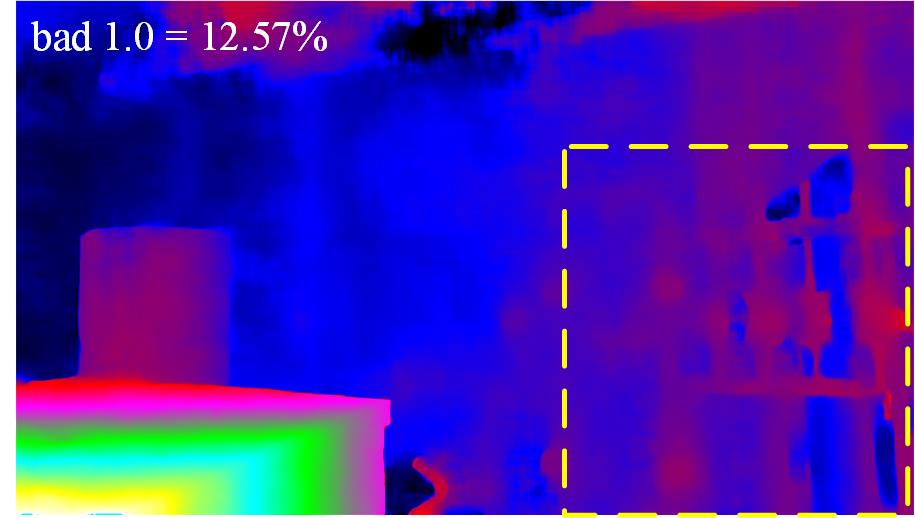}\\

	{(a) Left image} &  {(b) CFNet} &	{(c) GANet }	&  {(d) HSMNet }	  	\\
	\end{tabular}
	
	\caption{Visualization of some state-of-the-art methods’ performance on three real-world dataset testsets (from top to bottom: KITTI, Middlebury, and ETH3D). All methods are trained with a combination of KITTI, Middlebury, and ETH3D train images. GANet \cite{ganet} and HSMNet \cite{hsm} can achieve good performance on one specific dataset but perform poorly on the other two even if they have included targeted domain images in the training process. Our CFNet achieves SOTA or near SOTA performance on all three datasets without any adaptation.}
	\label{fig: introduction comparison}
\end{figure*}

However, current state-of-the-art methods are generally limited to a specific dataset due to the significant domain shifts across different datasets. For example, the KITTI dataset \cite{kitti1,kitti2} focuses on real-world urban driving scenarios while Middlebury \cite{mid} concentrates on indoor high-resolution scenes. Consequently, methods that are state-of-the-art on one dataset often cannot achieve comparable performance on a different one without substantial adaptation (visualization comparison can be seen in Fig.~\ref{fig: introduction comparison}). 
%%%   This is a general issue that occurs in many computer vision fields, such as image recognition \cite{robust_image_recognition} and depth prediction \cite{robust_depth_prediction}. 
However, real-world applications require the approaches to generalize well to different scenarios without adaptation. Thus, we need to push methods to be robust and perform well across different datasets with the fixed model parameters and hyperparameters.

\begin{figure}[!htb]
	\centering
% 	\tabcolsep=0.05cm
% 	\begin{tabular}{c c c }
%     \includegraphics[width=0.3\linewidth]{motivation_visual_final/disparity_distribution_kitti.png}&
%     \includegraphics[width=0.3\linewidth]{motivation_visual_final/disparity_distribution_middlebury.png}&
%     \includegraphics[width=0.3\linewidth]{motivation_visual_final/disparity_distribution_eth3d.png} \\
%    {(a) KITTI} &	{(b) Middlebury }	&  {(c) ETH3D }	  	\\
% 	\end{tabular}
    \includegraphics[width=0.6 \linewidth]{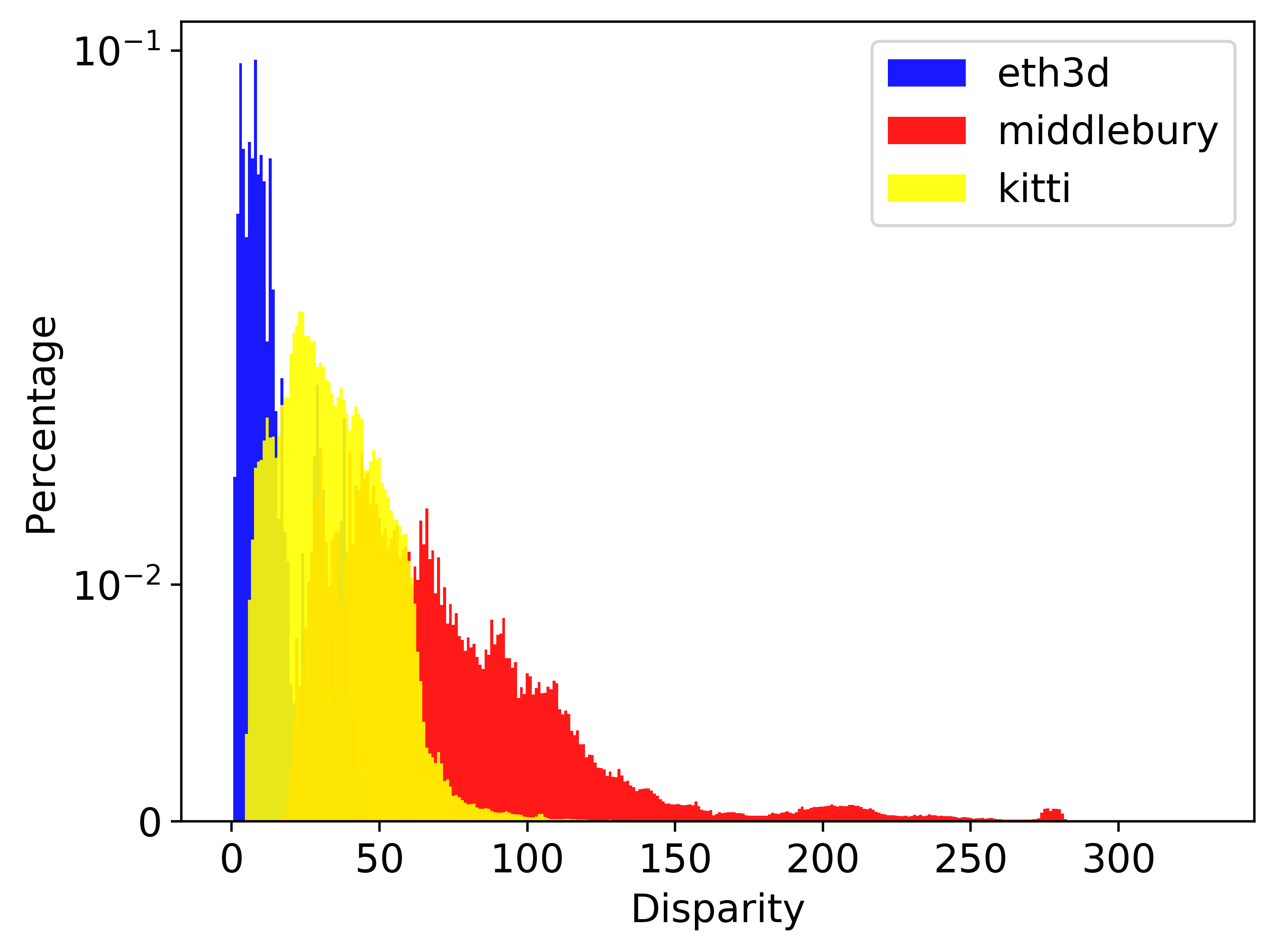} % requires the graphicx package
	\caption{Disparity distribution of KITTI 2015, Middlebury, and ETH3D training sets. We plot disparity distribution of half-resolution images for Middlebury while full-resolution images for other datasets. The disparity distribution across different datasets is unbalanced.}
	\label{fig: disparity distribution}
\end{figure}

The difficulties in designing a robust stereo matching system come from the large domain differences and unbalanced disparity distribution between a variety of datasets. As illustrated in Fig.~\ref{fig: introduction comparison} (a), there are significant domain differences across various datasets, \eg, indoors vs outdoors, color vs gray, and real vs synthetic, which leads to the learned features distorted and noisy \cite{dsmnet}. In addition, as illustrated in Fig.~\ref{fig: disparity distribution}, the disparity range of half-resolution images in Middlebury \cite{mid} is even more than 6 times larger than full-resolution images in ETH3D \cite{eth3d} (400 vs 64). Such unbalanced disparity distribution makes the current approaches trained with a fixed disparity range cannot cover the whole disparity range of another dataset without substantial adaption.

In this paper, we propose a cascade and fused cost volume representation to alleviate the above problems. 
(\textbf{1}) Towards the large domain differences, we propose to fuse multiple low-resolution dense cost volumes to enlarge the receptive field for capturing global and structural representations. Previous work \cite{dsmnet} observes that the limited effective receptive field of convolutional neural networks \cite{receptivefield} is the major reason the network is domain-sensitive to different datasets and proposes a learnable non-local layer to enlarge the receptive field. Inspired by it, we find that different scale low-resolution cost volumes can cover multi-scale receptive fields and are complementary to each other in promoting the network to look at different scale image regions. Thus, we can fuse multiple low-resolution dense cost volumes to guide the network to learn geometric scene information which is invariant across different datasets. In addition, this operation only adds a slight computation complexity.
(\textbf{2}) Towards the unbalanced disparity distribution, we propose a cascade cost volume representation and employ a variance-based uncertainty estimation to adaptively adjust the next stage disparity search range. That is, our method only needs to cover the union of all datasets’ disparity distribution (disparity range) at the first stage. Then we can employ our uncertainty estimation to evaluate pixel-level confidence of disparity estimation and prune out unlikely correspondences, guiding our network to look at more possible disparity search space at the next stage. In addition, we can save a lot of computational complexity by pruning out unlikely correspondences.

Experimentally, all methods are trained on the same training images and tested on three real-world datasets (KITTI2015, Middlebury, and ETH3D) with fixed model parameters and hyperparameters. As shown in Fig.~\ref{fig: robust comparison}, our method performs well on all three datasets and achieves state-of-the-art overall performance without adaptation.

In summary, our main contributions are:
\begin{itemize}
\item We propose a fused cost volume representation to reduce the domain differences across datasets.
\item We propose a cascade cost volume representation and develop a variance-based uncertainty estimation to balance different disparity distributions across datasets.
%\item We propose a cascade and fused cost volume representation to alleviate the large domain shifts and unbalanced disparity distribution between a variety of datasets.
\item Our method shows great generalization ability and obtains the 1st place on the stereo task of Robust Vision Challenge 2020.
\item Our method has great finetuning performance with low latency and ranks 1st on the popular KITTI 2015 and KITTI 2012 benchmarks among the published methods less than 200ms inference time. % faster than 200ms with adaptation on the target
\end{itemize}
%This is because our approach only needs to apply a shallow encoder-decoder architecture at a small resolution to predict an initial disparity map while DSMNet needs to repeatedly apply learnable no-local layers and 3D convolution layers at high resolution. 

\section{Related Work}
\subsection{Cost Volume based Deep Stereo Matching}
Stereo matching has been studied for decades and a well-known four-step pipeline \cite{scharstein2002taxonomy} has been established, where cost volume construction is an indispensable step. Typically, a cost volume is a 4D tensor of height, width, disparity, and features. Current state-of-the-art stereo matching methods are all cost volume based methods and they can be categorized into two categories. The first category uses a full correlation to generate a single-feature 3D cost volume. Such methods are usually efficient but lose much information due to of the decimation of feature channels. Many previous work, including Dispnet \cite{dispnet}, MADNet \cite{madnet}, and AANet \cite{aanet}, belong to this category. The second category usually uses concatenation \cite{gcnet} or group-wise correlation \cite{gwcnet} to generate a multi-feature 4D cost volume. Such methods can achieve improved performance while requiring higher computational complexity and memory consumption. Actually, a majority of the top-performing networks in public leaderboards belong to this category, such as GANet \cite{ganet} and CSPN \cite{cspn}. These methods generally employ multiple 3D convolution layers to constantly regularize the 4D cost volume and then apply softmax over the disparity dimension to produce a discrete disparity probability distribution. The final predicted disparity is obtained by soft argmin \cite{gcnet}, where the output is susceptible to multi-modal disparity probability distributions. To address this issue, ACFNet \cite{acfnet} directly supervises the cost volume with unimodal ground truth distributions. In contrast, we define an uncertainty estimation to quantify the degree to which the cost volume tends to be multi-modal distribution, higher implies the higher possibility of estimation error.
%softly weighting indices according to their probability, which is also

\subsection{Multi-scale Cost Volume based Deep Stereo Matching} 
Multi-scale cost volume was firstly applied in the single-feature 3D cost volume based network with the form of two-stage refinement \cite{mcvmfc} and pyramidal towers \cite{madnet,pwcnet}. Recently, cascade cost volume representation \cite{cvpmvsnet, cascade, uscnet} was proposed in multi-view stereo to alleviate the high computational complexity and memory consumption in employing 4D cost volumes. These methods generally predict an initial disparity at the coarsest resolution. Then, they narrow down the disparity search space and gradually refine the disparity. Recently, Casstereo \cite{cascade} extends such representation to stereo matching. It uniformly samples a pre-defined range to generate the next stage’s disparity search range. Instead, we employ uncertainty estimation to adaptively adjust the next stage pixel-level disparity search range and push the next stage's cost volume to be predominantly unimodal. Our method also shares similarities with UCSNet \cite{uscnet}, which constructs uncertainty-aware cost volume in multi-view stereo. However, it only focuses on dataset-specific performance. In addition, it generates the next stage search range with pre-defined parameters while we use learned parameters, which can better adapt to different datasets.

\subsection{Robust Stereo Matching}
Recently, researchers have shown an increased interest in robust stereo matching. These methods can be roughly categorized into two types. 1) Cross-domain Generalization: This category aims to improve the generalization of the network to unseen scenes. Towards this end, Jia \etal \cite{sungeneralizaiton} propose to incorporate scene geometry priors into an end-to-end network. Zhang \etal \cite{dsmnet} introduce a domain normalization and a trainable non-local graph-based filter to construct a domain-invariant stereo matching network. 2) Joint Generalization: This category aims to push the network to perform well on a variety of datasets with the same model parameter. MCV-MFC \cite{mcvmfc} introduces a two-stage finetuning scheme to achieve a good trade-off between generalization capability and fitting capability in a number of datasets. However, it doesn’t touch the inner difference between diverse datasets. To further address this problem, we propose a cascade and fused cost volume representation to alleviate the domain shifts and disparity distribution unbalance between a variety of datasets. Additionally, our method also performs well on cross-domain generalization, which further emphasizes the robustness of our network.

\begin{figure*}[!htb]
 \centering
 \includegraphics[width=0.7 \linewidth]{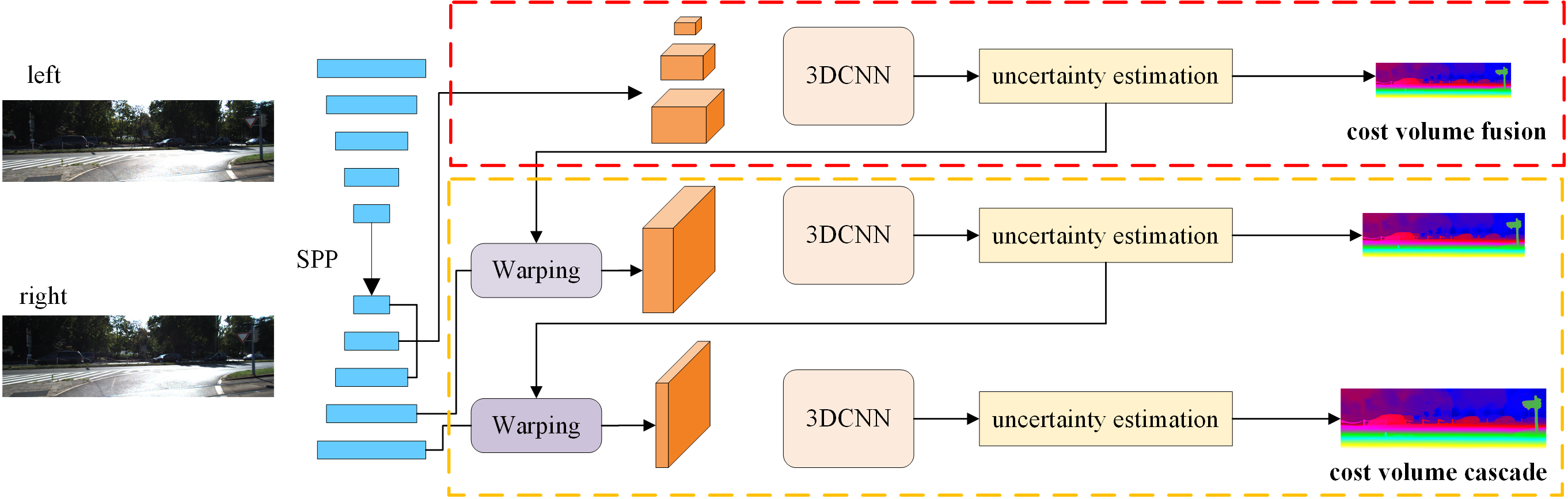} % requires the graphicx package
 \caption{The architecture of our proposed network. Our network consists of 3 parts: pyramid feature extraction, fused cost volume, and cascade cost volume.}
 \label{fig: architecture}
\end{figure*}

\section{Our Approach}
\subsection{Overview of CFNet}
We propose a cascade and fused cost volume representation for robust stereo matching. The overall architecture of our model is shown in Fig.~\ref{fig: architecture}, which consists of three parts: pyramid feature extraction, fused cost volume, and cascade cost volume. 

Given an image pair, we first employ a siamese unet-like \cite{hsm,unet} encoder-decoder architecture with skip connections to extract multi-scale image features. The encoder consists of five residual blocks, followed by an SPP module to better incorporate hierarchical context information. Our SPP module is similar to the one used in HSMNet \cite{hsm} while changing the size of average pooling blocks to $H/s \times W/s$, where $s \in \{32,64,96,128\}$. Compared with the widely used Resnet-like network \cite{cascade,gwcnet}, our method is more efficient and still contains sufficient information for cost aggregation. Experiments show that our pyramid feature extraction can achieve comparable performance with lower computational complexity. Then, we divide the multi-scale features into fused and cascade cost volume and predict multi-resolution disparity respectively.
% Firstly, we use both feature concatenation and group-wise correlation to generate lower-scale combination volumes and fuse them by an encoder-decoder process to get the initial disparity map. Then, an adaptive variance-based uncertainty estimation is employed to get the next stage disparity searching range, which can be used to generate the next stage’s combination volume. By iteratively narrowing down the disparity range and improving the cost volume resolution, the disparity estimation is gradually refined in a coarse-to-fine manner.
The details of cost volume fusion and cost volume cascade will be discussed in the next section.

\begin{figure}[!htb]
 \centering
 \includegraphics[width=0.85 \linewidth]{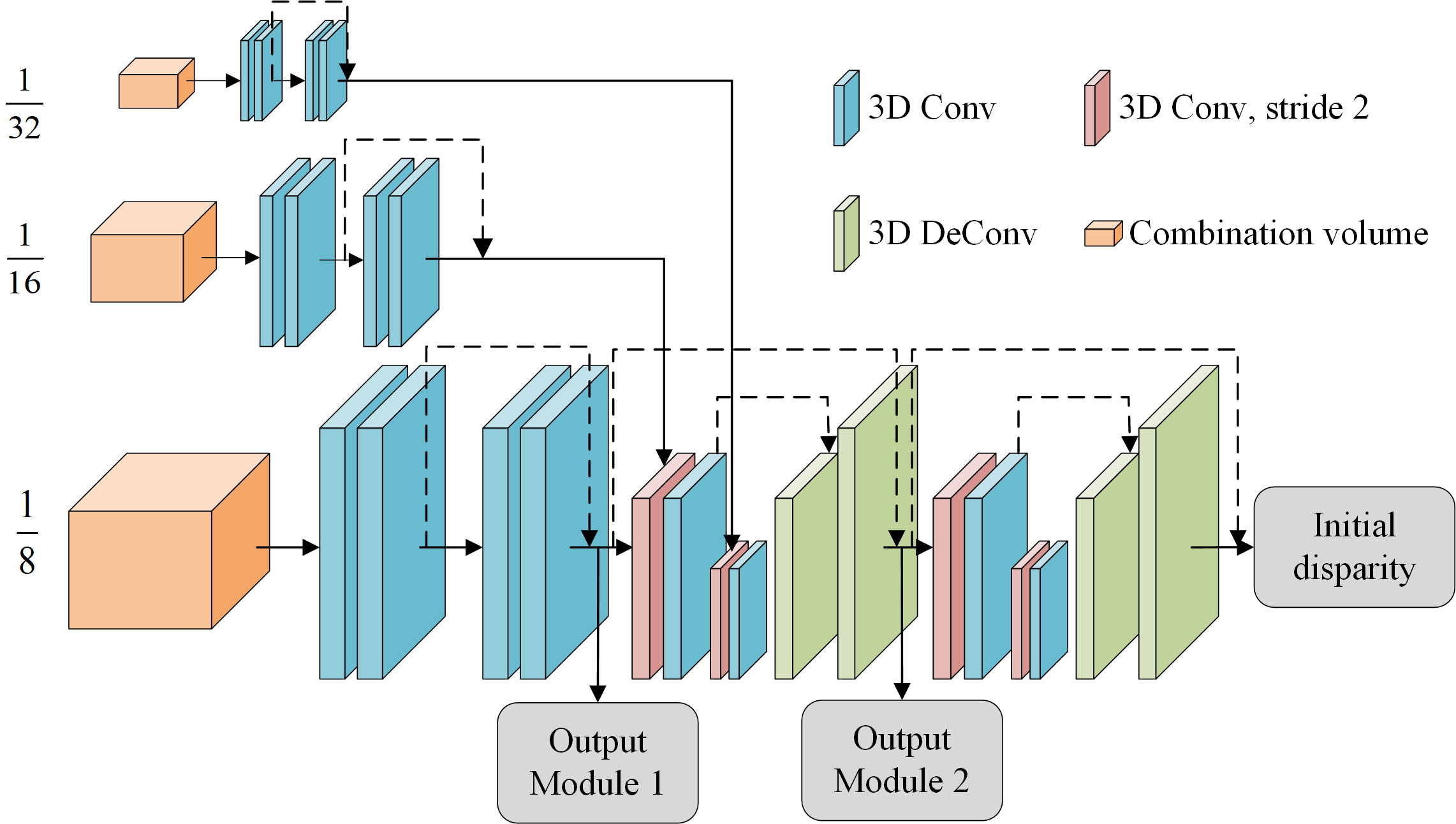} % requires the graphicx package
 \caption{The architecture of our cost volume fusion module. Three low-resolution cost volumes ($i \in (3,4,5)$) are fused to generate the initial disparity map. }
 \label{fig: fusion}
\end{figure}

\subsection{Fused Cost Volume}
We propose to fuse multiple low-resolution dense cost volumes (smaller than 1/4 of the original input image resolution in our paper) to reduce the domain shifts between different datasets for initial disparity estimation. Existing approaches \cite{cascade,uscnet,cvpmvsnet} have realized the importance of employing multi-scale cost volumes. However, these methods generally abandon small-resolution cost volumes because such cost volumes don’t contain sufficient information to generate an accurate disparity map respectively. Instead, we argue that different scale small-resolution cost volumes can cover multi-scale receptive fields, guiding the network to look at different scale image regions. Thus, they can be fused together to extract global and structural representations and generate a more accurate initial disparity map than higher-resolution sparse cost volume. More precisely, we first construct low-resolution cost volumes at each scale respectively, and then design a cost volume fusion module to integrate them in an encoder-decoder process. We provide details about these two steps below.

\textbf{Cost volume construction:} Inspired by \cite{msmdnet,gwcnet}, we propose to use both feature concatenation and group-wise correlation to generate combination volume. The combination volume is computed as: 
\begin{eqnarray}
\begin{array}{c}
V_{concat}^i({d^i},x,y,f) = f_L^i(x,y)||f_R^i(x - {d^i},y) \\
\\
V_{gwc}^i({d^i},x,y,g) = \frac{1}{{N_c^i/{N_g}}}\left\langle {f_l^{ig}(x,y),f_r^{ig}(x - {d^i},y)} \right\rangle \\
\\
V_{combine}^i = V_{concat}^i||V_{gwc}^i
\end{array}
\label{eq:volume construction}
\end{eqnarray}
where $||$ denotes the vector concatenation operation. ${N_c}$ represents the channels of extracted feature. ${N_g}$ is the amount of group. 
$\left\langle {{\rm{ }},{\rm{ }}} \right\rangle$ represents the inner product. ${f^i}$ denotes the extracted feature at scale (stage) $i$ and  $i = 0$ represents the original input image resolution. 

Note that the disparity searching index ${d^i}$ is defined as 
${d^i} \in \{ 0,1,2 \ldots \frac{{{D_{\max }}}}{{{2^i}}} - 1\}$ and the hypothesis plane interval equals to 1 in the fused cost volume representation. That is, these cost volumes are all dense cost volumes with the size of  $\frac{H}{{{2^i}}} \times \frac{W}{{{2^i}}} \times \frac{{{D_{\max }}}}{{{2^i}}} \times F$. By densely sampling the whole disparity range in small resolution, we can efficiently generate a coarsest disparity map. Then we can employ variance-based uncertainty estimation to narrow down the disparity searching space at higher resolution and refine the disparity estimation in a coarse-to-fine manner. Details will be introduced in Section 3.3.

\textbf{Cost Volume fusion:} Following the method proposed in \cite{msmdnet}, we use an improved encoder-decoder architecture to fuse low-resolution cost volumes. The architecture is shown in Fig.~\ref{fig: fusion}. Specifically, we first employ four 3D convolution layers with skip connections to regularize each cost volume and use a 3D convolution layer (stride of two) to down-sample the combination volume of scale 3 from 1/8 to 1/16 of the input image resolution. Next, we concatenate them (the down-sampled cost volume and the next stage combination volumes) at the feature dimension and then decrease the feature channel to a fixed size via one additional 3D convolution layer. Then, we apply a similar operation to progressively down-sample the cost volume to 1/32 of the original input image resolution and adopt 3-D transposed convolution to up-sample the volume in the decoder. In addition, we utilize one 3-D hourglass network to further regularize and refine the volume. Finally, an output module is applied to predict the disparity. The output module contains two more 3D convolution layers, aiming at obtaining a 1-channel 4D volume. To transform volume into disparity, we apply soft argmin \cite{gcnet} operation to generate initial disparity map ${D^3}$. The soft argmin operation is defined as:
\begin{eqnarray}
\widehat {{d^i}}{\rm{ = }}\sum\limits_{d = 0}^{\frac{{{D_{\max }}}}{{{2^i}}} - 1} {d \times \sigma ( - c_d^i)},
\end{eqnarray}
where $\sigma$ denotes the softmax operation and $c$ represents the predicted 1-channel 4D volume. ${\sigma ( - c_d)}$ denotes the discrete disparity probability distribution and the estimated disparity map is susceptible to all disparity indexes.

%\end{figure}
%\begin{figure}[!htb]
%	\centering
%	\tabcolsep=0.05cm
%	\begin{tabular}{c c}
%
%	\includegraphics[width=0.48\linewidth]{picture/ue_sample/example_unimodal.png}&
%    \includegraphics[width=0.48\linewidth]{picture/ue_sample/example_nearunimodal.png}\\
%
%    {     \scriptsize$\hat d = 2.0,U = 0.0$} &  {     \scriptsize$\hat d = 2.2,U = 0.16$} \\
%    {     \scriptsize(a)Unimodal distribution} &  {     \scriptsize(b)Predominantly unimodal distribution} \\
%
%	\includegraphics[width=0.48\linewidth]{picture/ue_sample/example_multimodal.png}&
%    \includegraphics[width=0.48\linewidth]{picture/ue_sample/example_error.png}\\
%    {     \scriptsize$\hat d = 2.7,U = 1.01$} &  {     \scriptsize $\hat d = 2.0,U = 2.0$} \\
%    {     \scriptsize(c) Multi-modal distribution1} &  {     \scriptsize(d)Multi-modal distribution2}
%
%	\end{tabular}
%	\caption{Some samples of uncertainty estimation. Expected value (ground truth) is 2px. The disparity searching range is from 0 to 4.}
%	\label{fig: uncertainty estimation sample}
%\end{figure}

\subsection{Cascade Cost Volume}
Given the initial disparity estimation, the next step is to construct a fine-grained cost volume and refine disparity maps in a coarse-to-fine manner. Considering the next stage disparity search range, uniform sampling a pre-defined range is the most straightforward way \cite{cascade}. However, such an approach assumes that all pixels are the same and cannot make adaptive pixel-level adjustments. For example, we should expand the searching range of ill-posed and occluded pixels. Furthermore, the disparity distribution of datasets with different characteristics is usually unbalanced. Thus, a question arises, can we make networks avoid being affected by invalid disparity indexes in a large initial disparity range and capture more possible pixel-level disparity search space with the prior knowledge of the last stage’s disparity estimation.

\begin{figure}[!htb]
	\centering
	\tabcolsep=0.05cm
	\begin{tabular}{c c c}

	\includegraphics[width=0.31\linewidth]{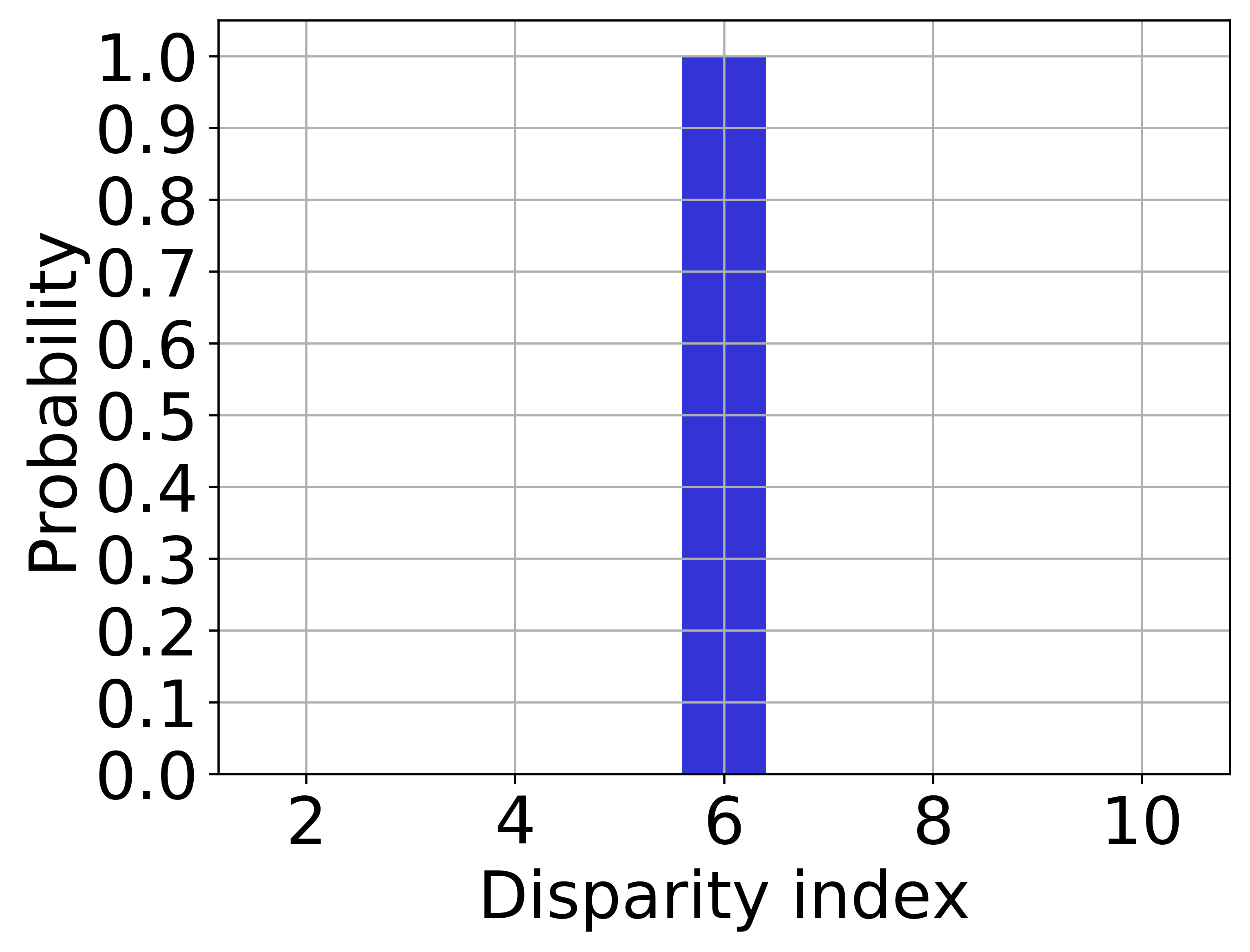}&
    \includegraphics[width=0.31\linewidth]{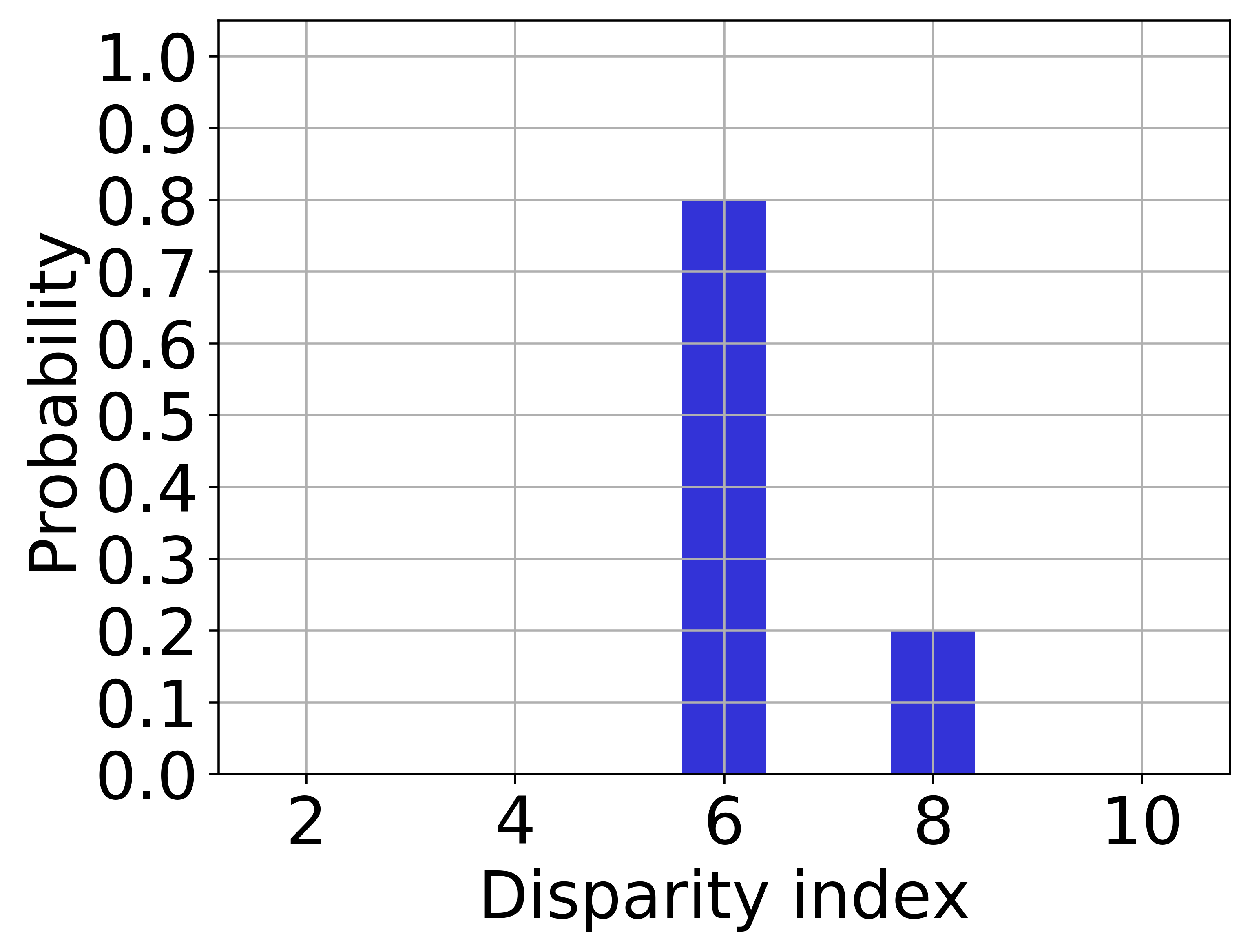}&
     \includegraphics[width=0.31\linewidth]{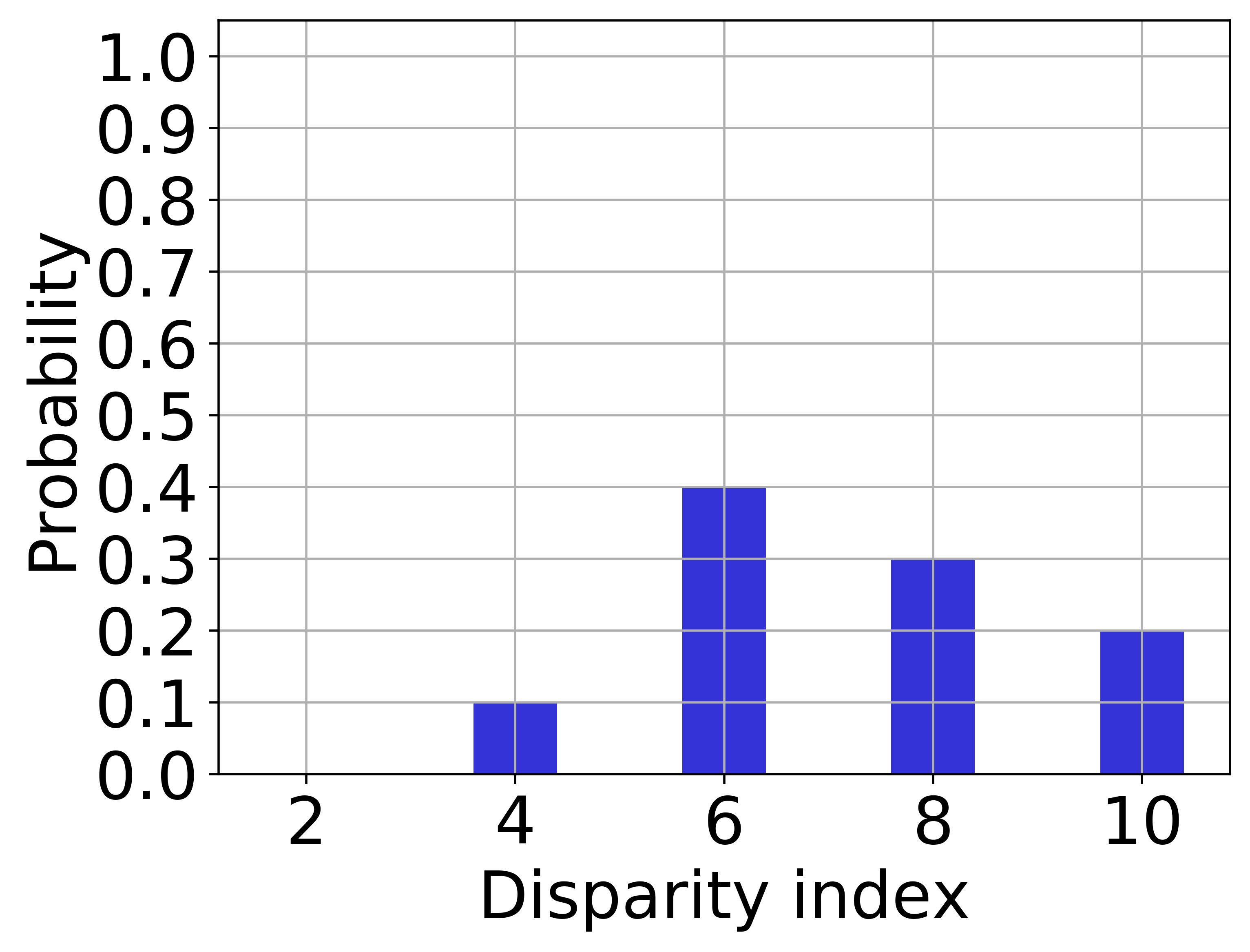}\\

    {     \scriptsize$\hat d = 6.0,U = 0.0$} &  {     \scriptsize$\hat d = 6.4,U = 0.64$} &  {     \scriptsize$\hat d = 7.2,U = 3.36$}\\
    {\scriptsize(a)Unimodal} &  {\scriptsize(b)Predominantly unimodal} & {\scriptsize(c)Multi-modal}

%	\includegraphics[width=0.48\linewidth]{picture/ue_sample/example_multimodal.png}&
%    \includegraphics[width=0.48\linewidth]{picture/ue_sample/example_error.png}\\
%    {     \scriptsize$\hat d = 2.7,U = 1.01$} &  {     \scriptsize $\hat d = 2.0,U = 2.0$} \\
%    {     \scriptsize(c) Multi-modal distribution1} &  {     \scriptsize(d)Multi-modal distribution2}

	\end{tabular}
	\caption{Some samples of uncertainty estimation. Expected value (ground truth) is 6px. The disparity searching range is from 2 to 10 with 5 hypothesis planes.}
	\label{fig: uncertainty estimation sample}
\end{figure}

To tackle this problem, we propose an adaptive variance-based disparity range uncertainty estimation. As introduced in related work, discrete disparity probability distribution reflects the similarities between candidate matching pixel pairs and the final predicted disparity is a weighted sum of all disparity indexes according to their probability. Thus, the ideal disparity probability distribution should be unimodal peaked at true disparities. However, the actual probability distribution is predominantly unimodal or even multi-modal at some pixels. Previous work \cite{acfnet,gcnet} has discovered that the degree of multimodal distribution is highly correlated with the probability of prediction error. In addition, ill-posed areas, texture-less regions, and occlusions tend to be multimodal distribution as well as high estimation error rate. Therefore, we propose to define an uncertainty estimation to quantify the degree to which the cost volume tends to be multi-modal distribution and employ it to evaluate the pixel-level confidence of the current estimation. The uncertainty is defined as:
\begin{eqnarray}
\begin{array}{c}
{{\rm{U}}^i}{\rm{ = }}\sum\limits_{\forall {d^i}} {{{(d - {{\hat d}^i})}^2} \times \sigma ( - c_d^i)}, \\
\\
\widehat {{d^i}}{\rm{ = }}\sum\limits_{\forall {d^i}} {d \times \sigma ( - c_d^i)}, 
\end{array}
\end{eqnarray}
where $\sigma$ denotes the softmax operation and $c$ represents the predicted 1-channel 4D volume. As shown in Fig.~\ref{fig: uncertainty estimation sample}, the uncertainty of unimodal distribution equals to 0 and the more the probability distribution is toward the multi-modal distribution, the higher the uncertainty and error.
%We also notice that some special multi-modal distribution (\autoref{fig: uncertainty estimation sample} (d)) can achieve accurate estimation with high uncertainty. However, such a situation rarely happens and most of the multi-modal distribution leads to inaccurate estimation \cite{acfnet}. 
Therefore, it’s reasonable to employ uncertainty to evaluate the confidence of disparity estimation, higher uncertainty implies a higher probability of prediction error and a wider disparity searching space to correct the wrong estimation (visualization can be seen in Fig.~\ref{fig: ue visual}). Thus, the next stage’s disparity searching range is defined as:
\begin{eqnarray}
\begin{array}{l}
d_{\max }^{i - 1} = \delta (\widehat {{d^i}} + \left( {{\alpha ^i} + 1} \right)\sqrt {{U^i}}  + {\beta ^i}),\\
d_{\min }^{i - 1} = \delta (\widehat {{d^i}} - \left( {{\alpha ^i} + 1} \right)\sqrt {{U^i}}  - {\beta ^i}),
\end{array}
\end{eqnarray}
where $\delta$ denotes bilinear interpolation. $\alpha$ and $\beta$ are normalization factors, which is initialized as 0 and gradually learns a weight. $\alpha$ and $\beta$ can also be set as hyper parameters, while experiment shows that the learned parameters are more robust than human-selected parameters \cite{uscnet}. Then we can employ uniform sampling to get next stage discrete hypothesis depth planes ${d^{i - 1}}$:
\begin{eqnarray}
\begin{array}{c}
{d^{i - 1}} = d_{\min }^{i - 1} + n(d_{\max }^{i - 1} - d_{\min }^{i - 1})/\left( {{N^{i - 1}} - 1} \right),\\
\\
n \in \{ 0,1,2 \ldots {N^{i-1}} - 1\},
\end{array}
\end{eqnarray}
where ${N^{i - 1}}$ is the number of hypothesis planes at stage $i-1$. Then, a fine-grained cost volume at stage $i-1$ is similarly defined based on Eq.\ref{eq:volume construction}, which leads to a sparse cost volume with the size of $\frac{H}{{{2^{i - 1}}}} \times \frac{W}{{{2^{i - 1}}}} \times {N^{i - 1}} \times F$. After getting the next stage cost volume, similar cost aggregation network (omitting the solid line in Figure \ref{fig: fusion}) can be employed to predict this stage’s disparity map. By iteratively narrow down the disparity range and higher the cost volume resolution, we can refine the disparity in a coarser to fine manner.

\begin{figure}[!htb]
	\centering
	\tabcolsep=0.05cm
	\begin{tabular}{c c}

	\includegraphics[width=0.43\linewidth]{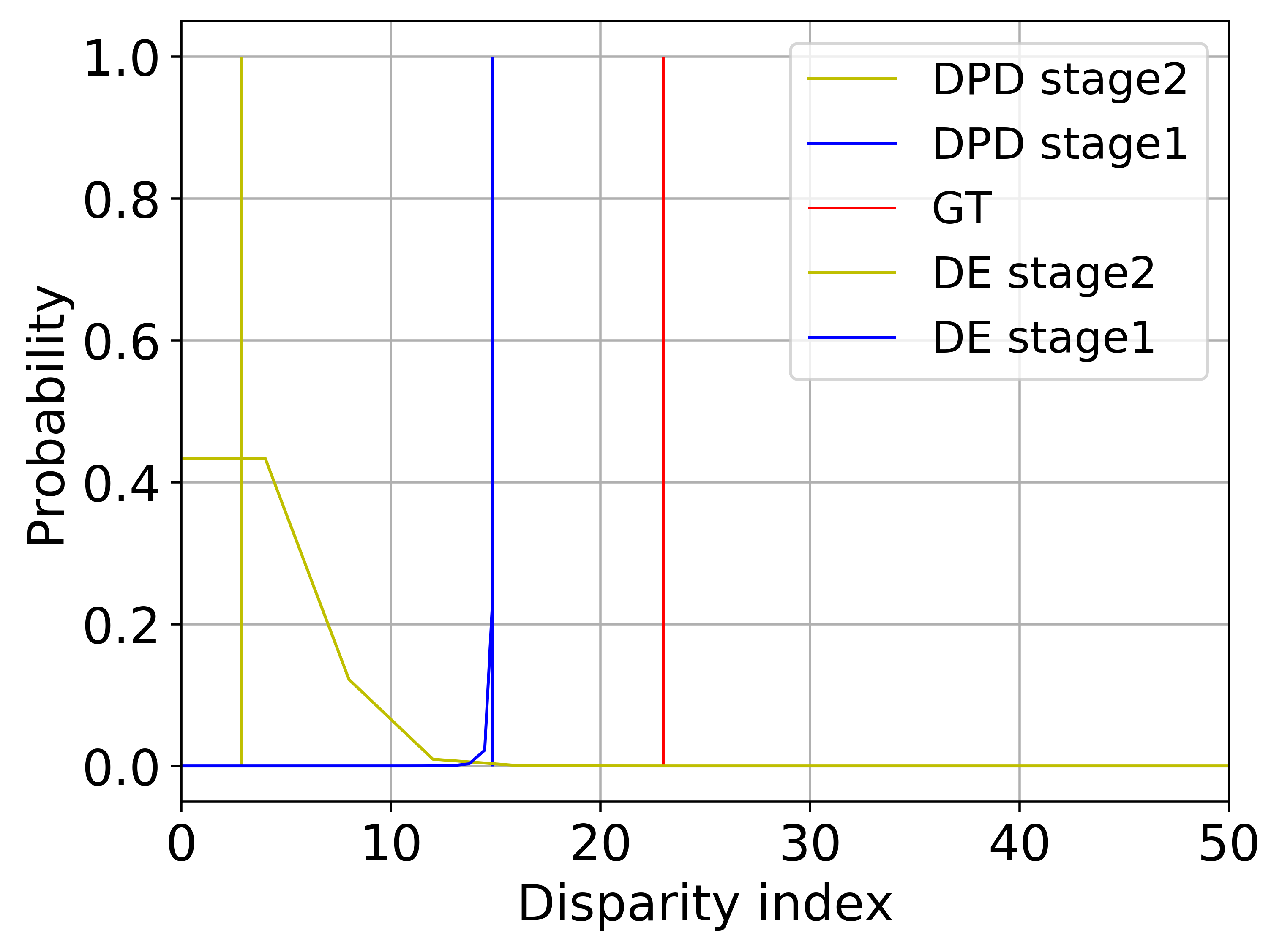}&
	\includegraphics[width=0.43\linewidth]{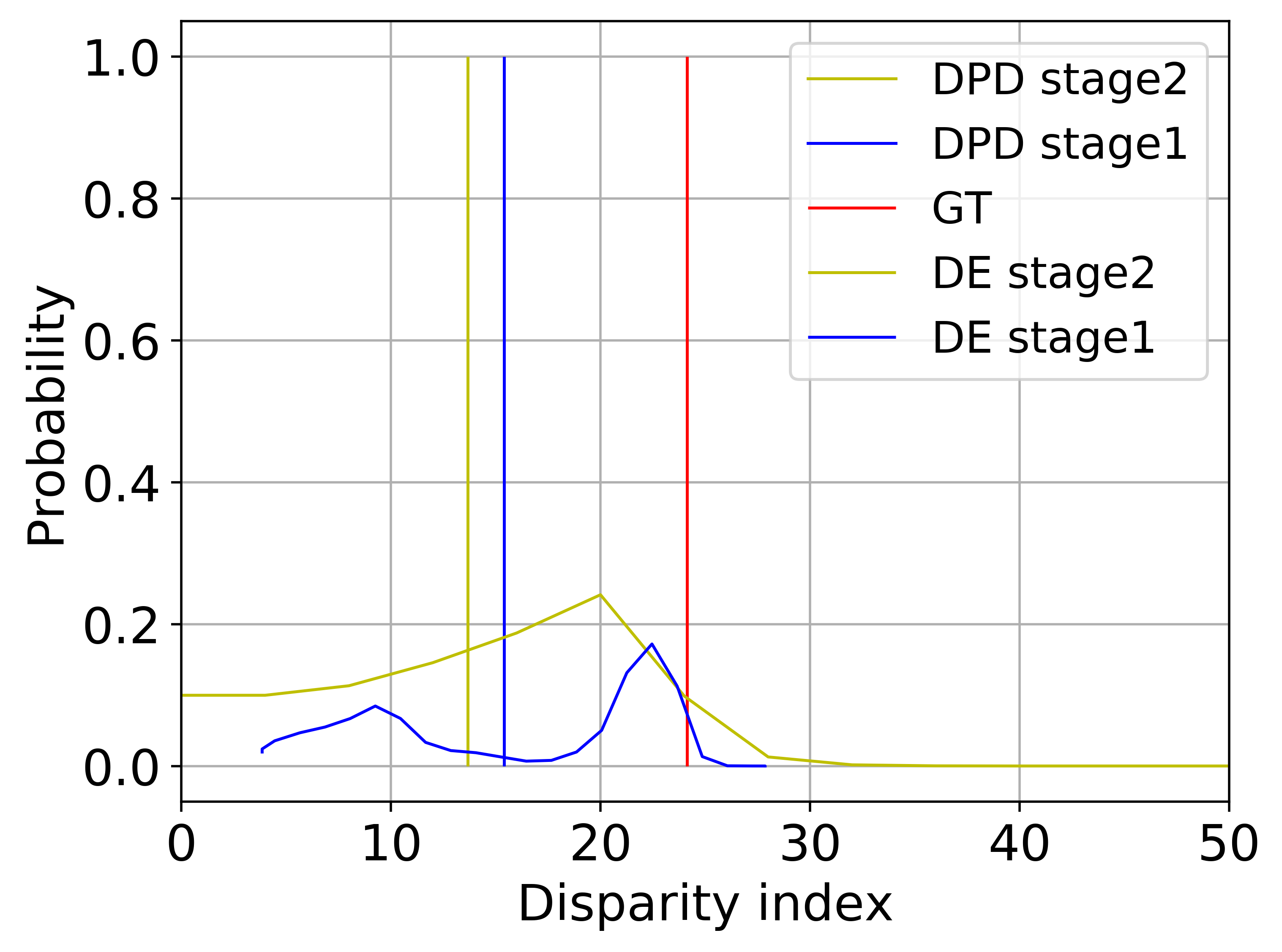}\\

	\includegraphics[width=0.43\linewidth]{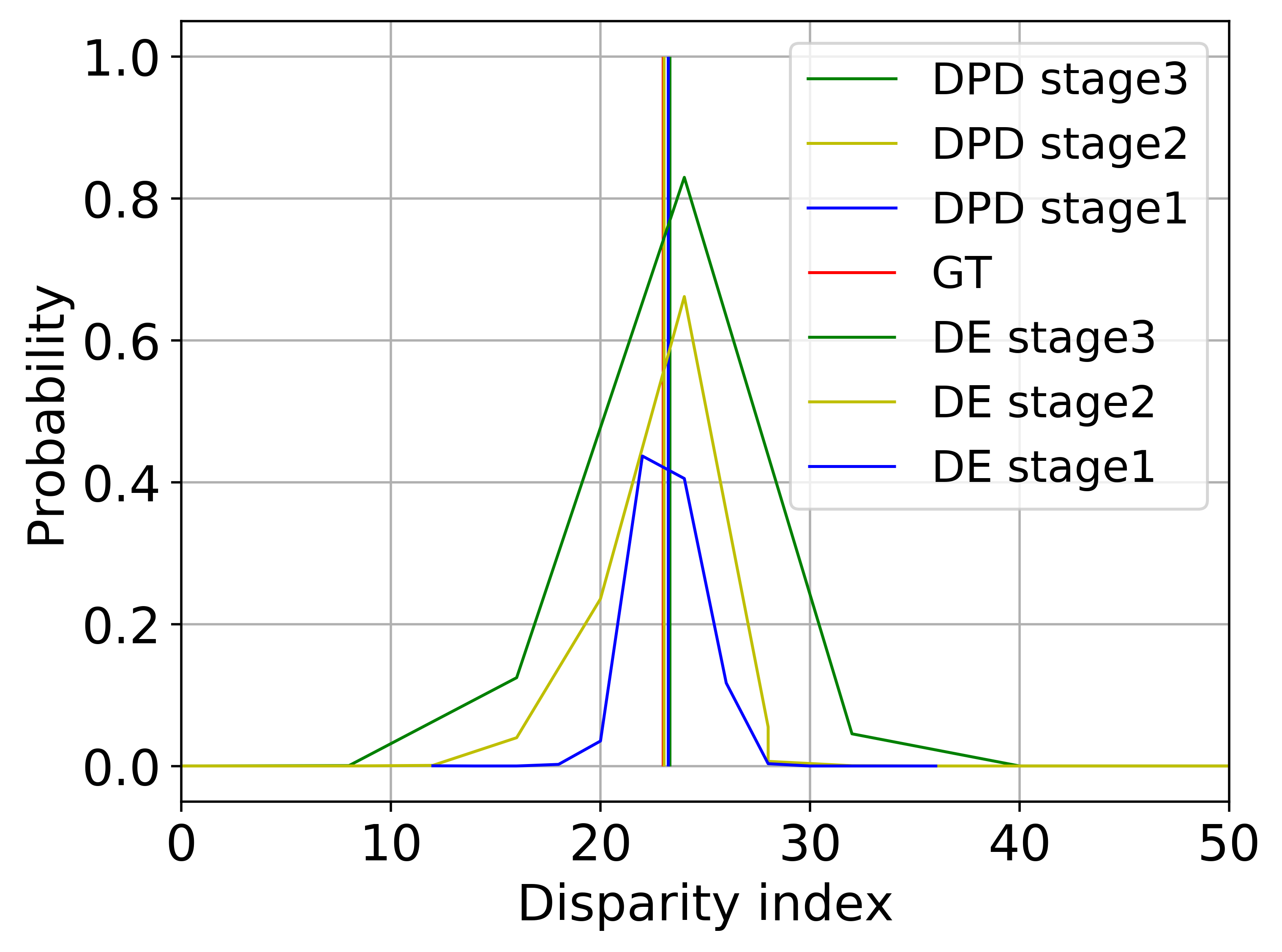}&
	\includegraphics[width=0.43\linewidth]{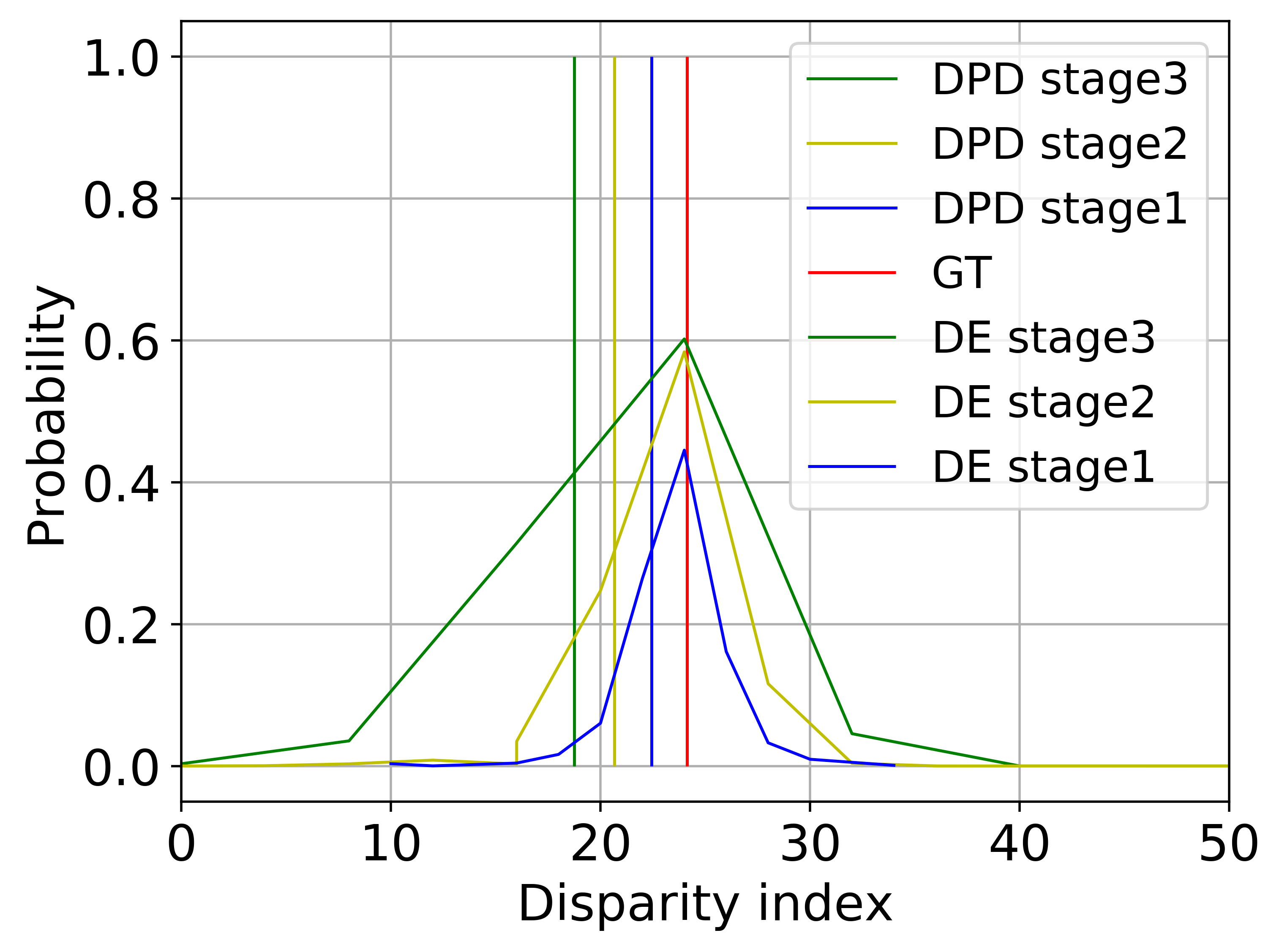}\\

    {(a)} &  {(b)}

	\end{tabular}

%	\caption{\small Visualization results on KITTI 2012 and KITTI 2015 testset. The left panel shows the left input image of stereo image pair, and for each example, the error maps of MSMD-Net, GWCNet and GANet-deep are presented.}
	\caption{Some examples of disparity probability distributions of cascade stereo (first row) and CFNet (second row). DPD: disparity probability distribution, DE: disparity estimation, GT: ground truth. Final disparity estimation is the estimation of stage 1.
%Our method can predict a more exact next stage disparity search range and a more reasonable predominantly unimodal distribution at each stage.
}

% The disparity estimation value is calculated by summing the product of each disparity’s probability and its disparity index. The ideal disparity probability distribution should be a unimodal distribution that peaked at the true disparity.
	\label{fig: cost volume distribution}
\end{figure}

While it is noteworthy that the insight of cascade cost volume has also been investigated in \cite{cascade}, our work differs from theirs in the following three main aspects: First, casstereo constructs higher-resolution sparse cost volume to predict an initial disparity estimation. Besides, previous work \cite{cvpmvsnet} argues that small resolution cost volume is too small to generate a reasonable initial disparity map to be refined. In contrast, we prove such cost volumes can even generate a more accurate initial disparity map than higher-resolution sparse cost volume with cost volume fusion. (As shown by the comparison between the estimation of casstereo at stage 2 and estimation of our CFNet at stage 3 in Fig.~\ref{fig: cost volume distribution} (a) and (b)). Second, casstereo only uniform sampling a pre-defined range to generate the next stage disparity search range. Instead, we develop a variance-based uncertainty estimation to adaptively adjust the next stage disparity search range which can push disparity distribution to be more predominantly unimodal (Fig.~\ref{fig: cost volume distribution}(b)). Third, by the cooperation of cascade and fused cost volume representation, our method can better cover the corresponding ground truth value in the final stage disparity search range and corrects some biased results in casstereo (Fig.~\ref{fig: cost volume distribution}(a)).

\section{Experiments}
% In this section, we discuss the datasets we used, data argumentation and implementation details. The datasets and data argumentation are described in Section 2.1 and 2.2. The implementation details are described in Section 2.3.  
\subsection{Dataset}
%\begin{itemize}

\noindent \textbf{SceneFlow:} This is a large synthetic dataset including 35,454 training and 4,370 test images with a resolution of $960\times 540$ for optical flow and stereo matching. We use it to pre-train our network. 

\noindent \textbf{Middlebury:} Middlebury \cite{mid} is an indoor dataset with 28 training image pairs (13 of them are additional training images) and 15 testing image pairs with full, half, and quarter resolutions. It has the highest resolution among the three datasets and the disparity range of half-resolution image pairs is 0-400. We select half-resolution image pairs to train and test our model.

\noindent \textbf{KITTI 2012\&2015:} They are both real-world datasets collected from a driving car. KITTI 2015 \cite{kitti2} contains 200 training and another 200 testing image pairs while KITTI 2012 \cite{kitti1} contains 194 training and another 195 testing image pairs. Both training image pairs provide sparse ground-truth disparity and the disparity range of them is 0-230.

\noindent \textbf{ETH3D:} ETH3D \cite{eth3d} is the only grayscale image dataset with both indoor and outdoor scenes. It contains 27 training and 20 testing image pairs with sparsely labeled ground-truth. It has the smallest disparity range among three datasets, which is just in the range of 0-64.

\subsection{Implementation Details}
We use PyTorch to implement our 3-stage network and employ Adam (${\beta _{\rm{1}}} = 0.9,{\beta _2} = 0.999$) to train the whole network in an end-to-end way. The batch size is set to 8 for training on 2 Tesla V100 GPUs and the whole disparity search range is fixed to 256 during the training and testing process. We employ the smooth L1 loss function to train our network and include all intermediate outputs in the loss weight. ${N^1}$ and ${N^2}$ are set as 12 and 16, respectively. Asymmetric chromatic augmentation and asymmetric occlusion \cite{hsm} are employed for data augmentation. 

% Specifically, we apply different chromatic augmentation to the image pairs hoping our network can improve its robustness when stereo cameras are under different lighting and exposure conditions. We also randomly replace a rectangle region at the target image with the RGB means of the whole picture so that our network can learn to predict the disparity without correspondences.

Inspired by the two-stage finetuning strategy \cite{mcvmfc}, we propose a three-stage finetune strategy to train our network. First, following the method proposed in \cite{msmdnet}, we use switch training strategy to pre-train our model in the SceneFlow dataset. Specifically, we first use ReLU to train our network from scratch for 20 epochs, then we switch the activation function to Mish and prolong the pre-training process in the SceneFlow dataset for another 15 epochs. Second, we jointly finetune our pre-train model on four datasets, i.e., KITTI 2015, KITTI2012, ETH3D, and Middlebury for 400 epochs. The initial learning rate is 0.001 and is down-scaled by 10 after epoch 300. Third, we augment Middlebury and ETH3D to the same size as KITTI 2015 and finetune our model for 50 epochs with a learning rate of 0.0001. The core idea of our three-stage finetune strategy is to prevent the small datasets from being overwhelmed by large datasets. 
By augmenting small datasets at stage three and train our model with a small learning rate, our strategy makes a better trade-off between generalization capability and fitting capability on three datasets.

\renewcommand\arraystretch{1.4}
\begin{table}[!htb]
\centering
\resizebox{0.48\textwidth}{!}{
\begin{tabular}{c|c|c|c|c|c}
\hline
Experiment                           & Method                           & \begin{tabular}[c]{@{}c@{}}KITTI\\ D1\_all\end{tabular} & \begin{tabular}[c]{@{}c@{}}Middlebury\\ bad 2.0\end{tabular} & \begin{tabular}[c]{@{}c@{}}ETH3D\\ bad 1.0\end{tabular} & \begin{tabular}[c]{@{}c@{}}time\\ (s)\end{tabular} \\ \hline
\multirow{2}{*}{Feature Extraction}  & Resnet-like-network              & 1.76                                                    & 22.81                                                         & \textbf{3.49}                                           & 0.270                                              \\  
                                     & \underline{Pyramid Feature Extraction} & \textbf{1.71}                                           & \textbf{22.27}                                               & 3.57                                                    & \textbf{0.225}                                     \\ \hline
\multirow{2}{*}{Cost Volume Fusion}  & Not Fuse                         & 1.79                                                    & 22.65                                                        & 3.67                                                    & \textbf{0.220}                                     \\  
                                     & \underline{Fuse}                       & \textbf{1.71}                                           & \textbf{22.27}                                               & \textbf{3.57}                                           & 0.225                                              \\ \hline
\multirow{3}{*}{Cost Volume Cascade} & Uniform Sample                   & 1.92                                                    & 23.8                                                         & 3.97                                                    & 0.225                                              \\  
                                     & UE + Hyperparameters             & 1.78                                                    & 23.13                                                        & 3.83                                                    & 0.225                                              \\  
                                     & \underline{UE + Learned Parameters}    & \textbf{1.71}                                           & \textbf{22.27}                                               & \textbf{3.57}                                           & \textbf{0.225}                                     \\ \hline
\multirow{3}{*}{Loss weight}         & Loss1:1,Loss2:1,Loss3:1          & 1.84                                                    & 23.87                                                        & \textbf{3.47}                                           & 0.225                                              \\ 
                                     & Loss1:1,Loss2:0.7,Loss3:0.5      & 1.79                                                    & 22.97                                                        & 3.62                                                    & 0.225                                              \\  
                                     & \underline{Loss1:2,Loss2:1,Loss3:0.5}  & \textbf{1.71}                                           & \textbf{22.27}                                               & 3.57                                                    & \textbf{0.225}                                     \\ \hline
\multirow{3}{*}{Fine-tuning strategy} & two stages               & \textbf{1.70}                                                    & 22.77                                                        & 3.99                                           & 0.234                                              \\ 
                                      & three stages\_no augment & \textbf{1.70}                                                    & 22.57                                                        & 3.92                                                    & 0.234                                              \\  
                                      & \underline {three stages}       & 1.71                                           & \textbf{22.27}                                               & \textbf{3.57}                                                    & \textbf{0.234}                                     \\ \hline
\end{tabular}
}
\caption{Ablation study results of the proposed network on KITTI 2015, Middlebury, and ETH3D validation set. UE: uncertainty estimation. Loss $i$: the loss weight at stage $i$. $j$  stages: $j$-stage finetune strategy. Three stages\_no augment: three stages without small dataset augment.We test a component of our method individually in each section of the table and the approach which is used in our final model is underlined. Time is measured on the KITTI dataset by a single Tesla V100 GPU.}
\label{tab: ablation study}
\end{table}

\renewcommand\arraystretch{1}

\subsection{Ablation Study}

We perform various ablation studies to show the effectiveness of each component in our network. We divide 20\% of the smallest dataset (5 images) from each real dataset (KITTI 2015, Middlebury, and ETH3D) as a validation set and use the rest of them as a training set to finetune our pretrain model. Results are shown in Table \ref{tab: ablation study}. Below we describe each component in more detail.

\noindent \textbf{Feature extraction:} We compare our pyramid feature extraction with the most widely used Resnet-like-network \cite{cascade,gwcnet}. As shown, our pyramid feature extraction can achieve similar performance with a faster speed, likely because the employing of small scale features is also helpful in feature extraction.

\noindent \textbf{Cost volume fusion:} We fuse three small-resolution cost volumes to generate the initial disparity map. Here, we test the impact when only a single volume is used. Cost volume fusion can achieve better performance with a slight additional computational cost. 

% In addition, previous work \cite{cvpmvsnet} argues that small resolution cost volume can’t generate a good initial depth map to be refined. In contrast, as shown in Figure \ref{fig: cost volume distribution}, we prove such cost volumes can even generate a more accurate initial disparity map than higher-resolution sparse cost volume with cost volume fusion.

\begin{figure}[!htb]
	\centering
	\tabcolsep=0.05cm
	\begin{tabular}{c c}
    \includegraphics[width=0.45\linewidth]{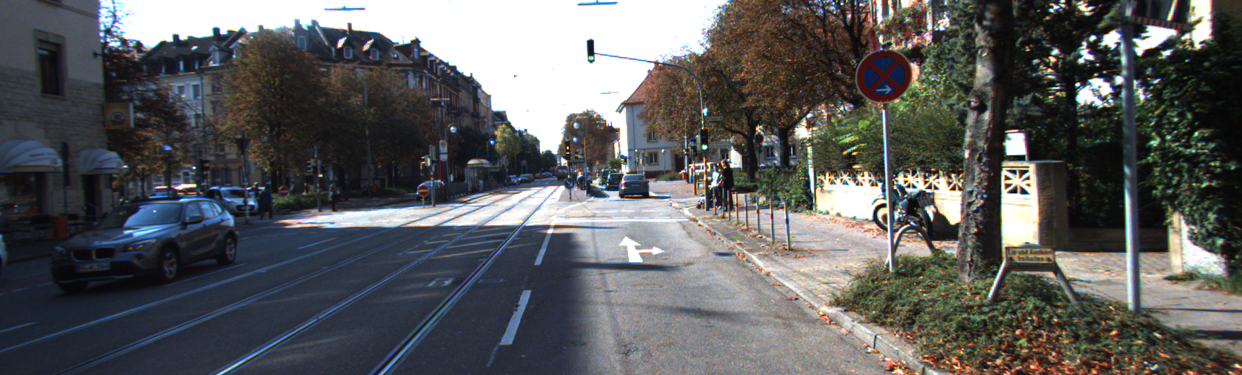}&
	\includegraphics[width=0.45\linewidth]{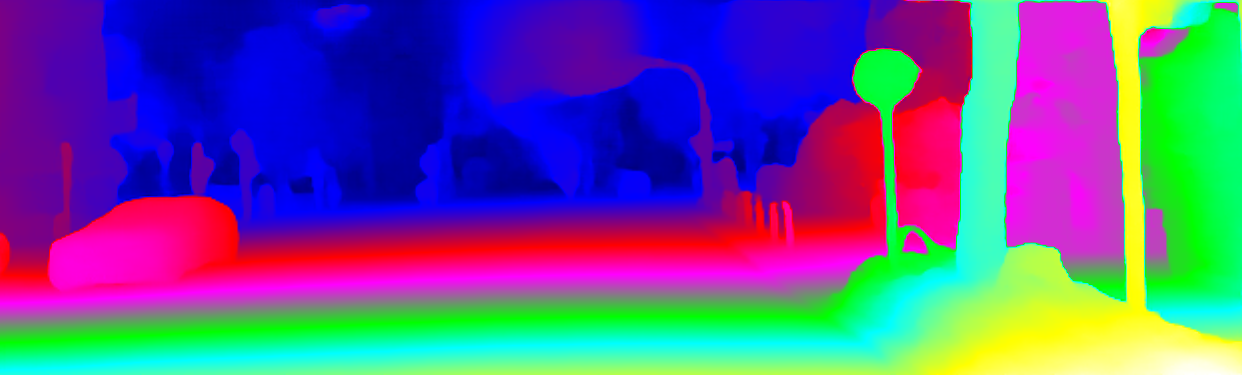}\\

    {(a) left image}   &  {(b) disparity map} \\
	
	\includegraphics[width=0.45\linewidth]{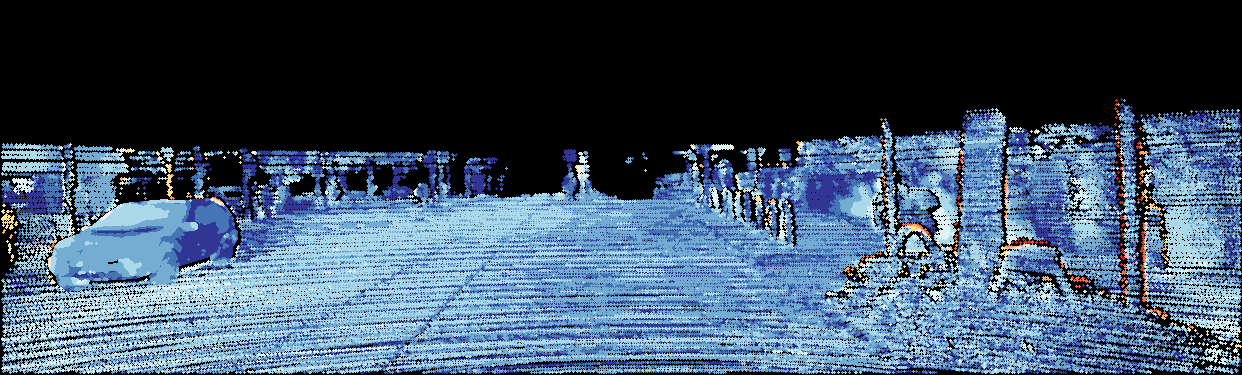}&
    \includegraphics[width=0.45\linewidth]{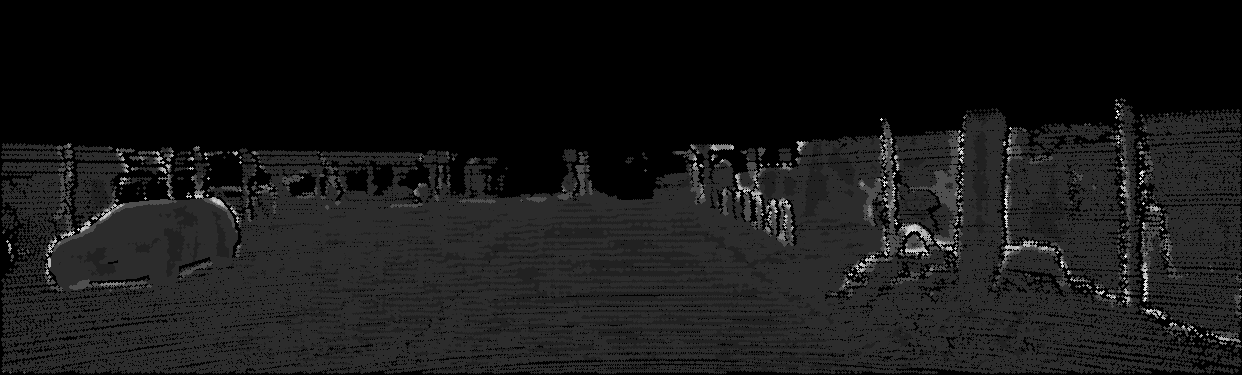}\\

	{(c) error map }	&  {(d) uncertainty map}	  	\\
	\end{tabular}
	
	\caption{Comparison between error map and uncertainty map. Red and white denote large errors and high uncertainty, respectively. As shown, the error map is highly correlated with the uncertainty map. See examples of the other two datasets in the supplementary material.}
	\label{fig: ue visual}
\end{figure}

\noindent \textbf{Cost volume cascade:} We test three ways of generating the next stage’s disparity searching space in cascade cost volume representation. As shown, learned parameters based uncertainty estimation achieves the best performance with tiny additional computation complexity. Furthermore, comparing with using hyperparameters, it is adaptive and eliminates the need to adjust hyperparameters according to different datasets. To further emphasize the effectiveness of our uncertainty estimation, we visualize the error map and uncertainty map in Fig.~\ref{fig: ue visual}. As shown, the error map is highly correlated with the uncertainty map. Experimentally, by removing 1\% of uncertain pixels ($\sqrt U  >  = 2.5$), we decrease the D1\_all error rate by 29.68\% (from 1.55\% to 1.09\% in KITTI 2015 validation set). Similar situations can be observed on the other two datasets. Thus, it is reasonable to employ uncertainty estimation to evaluate the pixel-level confidence of disparity estimation. 

\noindent \textbf{Loss weight:} The N-stages model outputs N disparity maps. We test three different settings of loss weight in our three-stage model. Experiments show that later stages shall set a larger loss weight which matches our intuition.

\noindent \textbf{Finetuning strategy:} We test three terms of finetuning strategy. As shown,  only extending the number of iterations can not improve the accuracy of predictions on small datasets. Instead, our strategy can greatly alleviate the problem of small datasets being overwhelmed by large ones.

\begin{table*}[!htb]
\centering
\resizebox{0.75\textwidth}{!}{
\begin{tabular}{c|c|c|c|c|c|c|c|c|c|c|c|c|c}
\hline
\multirow{2}{*}{Method}   & \multicolumn{4}{c|}{KITTI}                                 & \multicolumn{4}{c|}{Middlebury}                            & \multicolumn{4}{c|}{ETH3D}                                 & \multirow{2}{*}{\begin{tabular}[c]{@{}c@{}}Overall\\ Rank\end{tabular}} \\ \cline{2-13}
                          & D1\_bg        & D1\_fg        & D1\_all       & Rank & bad 1.0       & bad 2.0       & avg error     & Rank & bad 1.0       & bad 2.0       & avg error     & Rank &                                  \\ \hline
NLCANet\_V2\_RVC \cite{nlcanet}         & \textbf{1.51} & 3.97          & \textbf{1.92} & \textbf{1} & 29.4          & 16.4          & 5.60          & 3          & 4.11          & 1.2           & 0.29          & 2          & 2                                \\ 
HSMNet\_RVC \cite{hsm}               & 2.74          & 8.73          & 3.74          & 6          & 31.2          & 16.5          & \textbf{3.44} & \textbf{1} & 4.40          & 1.51          & 0.28          & 3          & 3                                \\ 
CVANet\_RVC               & 1.74          & 4.98          & 2.28          & 3          & 58.5          & 38.5          & 8.64          & 5          & 4.68          & 1.37          & 0.34          & 4          & 4                                \\ 
AANet\_RVC \cite{aanet}                & 2.23          & 4.89          & 2.67          & 5          & 42.9          & 31.8          & 12.8          & 6          & 5.41          & 1.95          & 0.33          & 5          & 5                                \\ 
GANet\_RVC \cite{ganet}                & 1.88          & 4.58          & 2.33          & 4          & 43.1          & 24.9          & 15.8          & 7          & 6.97          & 1.25          & 0.45          & 6          & 6                                \\ 
\textbf{CFNet\_RVC(ours)} & 1.65          & \textbf{3.53} & 1.96          & 2          & \textbf{26.2} & \textbf{16.1} & 5.07          & 2          & \textbf{3.7}  & \textbf{0.97} & \textbf{0.26} & \textbf{1} & \textbf{1}                       \\ \hline
iResNet\_ROB \cite{iresnet, mcvmfc}                   & 2.27          & 4.89          & 2.71          & 3          & 45.9          & 31.7          & 6.56          & 2          & 4.67          & 1.22          & 0.27          & 3          & 3                                \\ 
Deeppruner\_ROB \cite{deeppruner}                & -             & -             & 2.23          & 2          & 57.1          & 36.4          & 6.56          & 3          & 3.82          & 1.04          & 0.28          & 2          & 2                                \\ 
\textbf{CFNet\_RVC(ours)} & \textbf{1.65} & \textbf{3.53} & \textbf{1.96} & \textbf{1}          & \textbf{26.2} & \textbf{16.1} & \textbf{5.07} & \textbf{1}          &\textbf{3.7}           & \textbf{0.97}          & \textbf{0.26}          & \textbf{1}          & \textbf{1 }                               \\ \hline
\end{tabular}
}
\caption{Joint generalization comparison on ETH3D, Middlebury, and KITTI2015 datasets. \textbf{Top:} Generalization comparison with methods who participated in the Robust Vision challenge 2020. \textbf{Bottom:} Generalization comparison with the top 2 methods in the past two years. All methods are tested on three datasets without adaptation. The overall rank is obtained by Schulze Proportional Ranking \cite{rank} to joining multiple rankings into one. As shown, our method achieves the best overall performance.
% Note that other than typically ranking the performance with a single metric, to construct a robust system, every sub ranking is decided by a number of metrics per dataset.
}
\label{tab: robust challenge}
\end{table*}

\begin{table}[!htb]
\centering
\resizebox{0.42\textwidth}{!}{
\begin{tabular}{c|c|c|c|c}
\hline
Method    & \begin{tabular}[c]{@{}c@{}}KITTI2012\\ D1\_all(\%)\end{tabular} & \begin{tabular}[c]{@{}c@{}}KITTI2015\\ D1\_all(\%)\end{tabular} & \begin{tabular}[c]{@{}c@{}}Middlebury\\ bad 2.0(\%)\end{tabular} & \begin{tabular}[c]{@{}c@{}}ETH3D\\ bad 1.0(\%)\end{tabular} \\ \hline
PSMNet \cite{psmnet}   & 15.1                                                            & 16.3                                                            & 39.5                                                             & 23.8                                                        \\ 
GWCNet \cite{gwcnet}    & 12.0                                                            & 12.2                                                            & 37.4                                                             & 11.0                                                        \\ 
CasStereo \cite{cascade} & 11.8                                                            & 11.9                                                            & 40.6                                                             & 7.8                                                         \\ 
GANet \cite{ganet}    & 10.1                                                            & 11.7                                                            & 32.2                                                             & 14.1                                                        \\ 
DSMNet \cite{dsmnet}   & 6.2                                                             & 6.5                                                             & \textbf{21.8}                                                    & 6.2                                                         \\ \hline
CFNet     & \textbf{4.7}                                                    & \textbf{5.8}                                                    & 28.2                                                             & \textbf{5.8}                                                \\ \hline
\end{tabular}
}
\caption{Cross-domain generalization evaluation on ETH3D, Middlebury, and KITTI training sets. All methods are only trained on the Scene Flow datatest and tested on full-resolution training images of three real datasets.}
\label{tab: cross-domain generalization}
\end{table}

%\begin{figure*}[!htb]
%	\centering
%	\tabcolsep=0.05cm
%	\begin{tabular}{c c c c}
%    \includegraphics[width=0.24\linewidth]{ue_visual/000001_10.png}&
%	\includegraphics[width=0.24\linewidth]{ue_visual/000001_10_disp_s4_CF.png}&
%	\includegraphics[width=0.24\linewidth]{ue_visual/000001_10_disp_s3_CF.png}&
%	\includegraphics[width=0.24\linewidth]{ue_visual/000001_10_disp_s2_CF.png}\\
%	
%     &
%    \includegraphics[width=0.24\linewidth]{ue_visual/000001_10_error_s4_CF.png}&
%	\includegraphics[width=0.24\linewidth]{ue_visual/000001_10_error_s3_CF.png}&
%	\includegraphics[width=0.24\linewidth]{ue_visual/000001_10_error_s2_CF.png}\\
%
%    &
%    \includegraphics[width=0.24\linewidth]{ue_visual/000001_10_U_s4_CF}&
%	\includegraphics[width=0.24\linewidth]{ue_visual/000001_10_U_s3_CF}&
%	\includegraphics[width=0.24\linewidth]{ue_visual/000001_10_U_s2_CF}\\
%
%	{(a) left image} &  {(b) stage 3} &	{(c) stage 2 }	&  {(d) stage 1}	  	\\
%	\end{tabular}
%	
%	\caption{Visualization results of each stage. From top to bottom: disparity map, error map, and uncertainty (confidence) map (red and white denote large errors and high uncertainty, respectively). As shown, the error map is highly correlated with the uncertainty map. And we can find the disparity map is gradually refined as we higher the resolution. See examples of the other two datasets in the supplementary material. }
%	\label{fig: ue visual}
%\end{figure*}

\subsection{Robustness Evaluation}

As mentioned before, we define robustness as joint generalization. Such generalization is essential for current methods, which are limited to specific domains and cannot get comparable results on other datasets. This is also the goal of Robust Vision Challenge 2020\footnote[2]{\url{http://www.robustvision.net/}}. Towards this end, we evaluate methods' robustness by their performance on three real datasets without finetuning. 

We list some state-of-the-art methods’ performance in Table \ref{tab: robust challenge}. It can be seen from this table that HSMNet\_RVC \cite{hsm} ranks first on the Middlebury dataset. But it can’t get comparable results on the other two datasets (3rd on ETH3D and 6th on KITTI 2015). In particular, its performance on KITTI 2015 dataset is far worse than the other five. This is because this method is specially designed for high-resolution datasets and can’t generalize well on other datasets. GANet \cite{ganet} is the top-performing  method in the KITTI dataset. However, the error rate of D1\_all increased by 28.73\% (from 1.81\% to 2.33\%) after adding training images of the other two datasets and only ranks 4th on KITTI 2015. In addition, it still cannot get a good result on the other two datasets (6th on ETH3D and 7th on Middlebury). The similar situation also appeared on other methods. In contrast, our method shows great generalization ability and performs well on all three datasets (2nd on KITTI 2015, 1st on ETH3D, and 2nd on Middlebury) and achieves the best overall performance. We also compare our methods with the top two methods in the previous Robust Vision Challenge. As shown in Table \ref{tab: robust challenge}, our approach outperforms Deeppruner\_ROB and iResNet\_ROB on all three datasets with a remarkable margin.

We also notice that some previous work defines robustness as cross-domain generalization. To further emphasize the effectiveness of our method, we compare our method with some state-of-the-art methods by training on synthetic images and testing on real images. As shown in Table \ref{tab: cross-domain generalization}, our method far outperforms domain-specific methods \cite{psmnet,gwcnet,cascade,ganet} on all four datasets. DSMnet \cite{dsmnet} is specially designed for cross-domain generalization. Our method can surpass it on three datasets, which further shows our cascade and fused cost volume representation is an efficient approach for robust stereo matching.

\subsection{Results on KITTI Benchmark}
Although our focus is not on domain-specific performance, we still fine-tune our model on KITTI 2012 and KITTI 2015 benchmark to show the efficiency of our method. Specifically, we adjust each stage’s (except stage 3) stack hourglass number to be one and pretrain our model on Scene Flow datasets with the same training strategy. Then the pre-train model is finetuned on KITTI 2015 and KITTI 2012 datasets for 300 epochs, respectively. The learning rate starts at 0.001 and decreases to 0.0001 after 200 epochs. Following \cite{deeppruner}, we combine KITTI 2012 and KITTI 2015 image pairs for the evaluation of KITTI 2015 while only use KITTI 2012 image pairs for the evaluation of KITTI 2012.
 
Some SOTA real-time methods and best-performing approaches are listed in Table \ref{tab: KITTI benchmark}. We find that our method achieves 1.88\% D1\_all error rate, a 6\% error reduction from our base model casstereo \cite{cascade} with 3 times faster speed and gets similar performance with other best-performing approaches such as GANet-deep \cite{ganet} and ACFNet \cite{acfnet}. 
%Note that GANet-deep, AANet, Deeppruner, and HSMNet all include in the generalization evaluation and cannot achieve a good generalization. 
Furthermore, our method outperforms all published methods faster than 200ms with a noteworthy margin on both datasets which implies the efficiency of our method.

\begin{table}[!htb]
\centering
\resizebox{0.3\textwidth}{!}{
\begin{tabular}{c|c|c|c|c|c}
\hline
\multirow{2}{*}{Method} & \multicolumn{2}{c|}{\begin{tabular}[c]{@{}c@{}}KITTI2012\\ 3px(\%)\end{tabular}} & \multicolumn{2}{c|}{\begin{tabular}[c]{@{}c@{}}KITTI2015\\ D1\_all(\%)\end{tabular}} & \multirow{2}{*}{\begin{tabular}[c]{@{}c@{}}time\\  (s)\end{tabular}} \\ \cline{2-5}
                        & Noc                                     & All                                    & Noc                                       & All                                      &                                                                      \\ \hline
LEAStereo   \cite{lea}            & \textbf{1.13}                           & \textbf{1.45}                          & \textbf{1.51}                             & \textbf{1.65}                            & 0.3                                                                  \\ 
GANet-deep \cite{ganet}             & 1.19                                    & 1.60                                   & 1.63                                      & 1.81                                     & 1.8                                                                  \\ 
AcfNet \cite{acfnet}                 & 1.17                                    & 1.54                                   & 1.72                                      & 1.89                                     & 0.48                                                                 \\ 
Casstereo \cite{cascade}                & -                                       & -                                      & 1.78                                      & 2.0                                      & 0.6                                                                  \\ \hline
HITNet \cite{hitnet}                  & 1.41                                    & 1.89                                   & 1.74                                      & 1.98                                     & \textbf{0.015}                                                       \\ 
HD\textasciicircum{}3 \cite{hd3}   & 1.40                                    & 1.80                                   & 1.87                                      & 2.02                                     & 0.14                                                                 \\ 
AANet+ \cite{aanet}                  & 1.55                                    & 2.04                                   & 1.85                                      & 2.03                                     & 0.06                                                                 \\ 
HSMNet \cite{hsm}                  & 1.53                                    & 1.99                                   & 1.92                                      & 2.14                                     & 0.14                                                                 \\ 
Deeppruner \cite{deeppruner}              & -                                       & -                                      & 1.95                                      & 2.15                                     & 0.18                                                                 \\ 
\textbf{CFNet(ours)}    & \textbf{1.23}                           & \textbf{1.58}                          &  \textbf{1.73}                                        & \textbf{1.88}                            & 0.18                                                                 \\ \hline
\end{tabular}
}
\caption{Results on KITTI benchmark. \textbf{Top:} Comparison with best-performing methods. \textbf{Bottom:} Comparison with real-time methods. All methods are finetuned on specific datasets.}
\label{tab: KITTI benchmark}
\end{table}

\section{Conclusion}
We have proposed a cascade and fused cost volume representation for robust stereo matching. We first introduce a fused cost volume to alleviate the domain shifts across different datasets for initial disparity estimation. Then we construct cascade cost volume to balance the different disparity distribution across datasets, where the variance-based uncertainty estimation is at the core. We use it to adaptively narrow down the next stage’s pixel-level disparity searching space. Experiment results show that our approach performs well on a variety of datasets with high efficiency. In the future, we plan to extend our cost volume representation to semi-supervised or self-supervised setup \cite{self-supervised,Semi-supervised,self-supervised-stereo}.

\noindent\textbf{Acknowledgements} {\small This research was supported in part by National Key Research and Development Program of China (2018AAA0102803) and National Natural Science Foundation of China (61871325). We would like to thank the anonymous area chairs and reviewers for their useful feedback.}

{\small
\bibliographystyle{ieee_fullname}
\bibliography{egbib}

\begin{thebibliography}{10}\itemsep=-1pt

\bibitem{roboticsnavigation}
Joydeep Biswas and Manuela Veloso.
\newblock Depth camera based localization and navigation for indoor mobile
  robots.
\newblock In {\em RGB-D Workshop at RSS}, volume 2011, page~21, 2011.

\bibitem{psmnet}
Jia-Ren Chang and Yong-Sheng Chen.
\newblock Pyramid stereo matching network.
\newblock In {\em IEEE Conference on Computer Vision and Pattern Recognition
  (CVPR)}, pages 5410--5418, 2018.

\bibitem{autonomousdriving}
Chenyi Chen, Ari Seff, Alain Kornhauser, and Jianxiong Xiao.
\newblock Deepdriving: Learning affordance for direct perception in autonomous
  driving.
\newblock In {\em IEEE International Conference on Computer Vision (ICCV)},
  pages 2722--2730, 2015.

\bibitem{uscnet}
Shuo Cheng, Zexiang Xu, Shilin Zhu, Zhuwen Li, Li~Erran Li, Ravi Ramamoorthi,
  and Hao Su.
\newblock Deep stereo using adaptive thin volume representation with
  uncertainty awareness.
\newblock In {\em IEEE Conference on Computer Vision and Pattern Recognition
  (CVPR)}, pages 2524--2534, 2020.

\bibitem{cspn}
Xinjing Cheng, Peng Wang, and Ruigang Yang.
\newblock Learning depth with convolutional spatial propagation network.
\newblock In {\em IEEE Transactions on pattern analysis and machine
  intelligence (TPAMI)}, 2019.

\bibitem{lea}
Xuelian Cheng, Yiran Zhong, Mehrtash Harandi, Yuchao Dai, Xiaojun Chang,
  Hongdong Li, Tom Drummond, and Zongyuan Ge.
\newblock Hierarchical neural architecture search for deep stereo matching.
\newblock {\em Advances in Neural Information Processing Systems (NIPS)}, 33,
  2020.

\bibitem{deeppruner}
Shivam Duggal, Shenlong Wang, Wei-Chiu Ma, Rui Hu, and Raquel Urtasun.
\newblock Deeppruner: Learning efficient stereo matching via differentiable
  patchmatch.
\newblock In {\em IEEE International Conference on Computer Vision (ICCV)},
  pages 4384--4393, 2019.

\bibitem{slam1}
Jakob Engel, J{\"o}rg St{\"u}ckler, and Daniel Cremers.
\newblock Large-scale direct slam with stereo cameras.
\newblock In {\em IEEE International Conference on Intelligent Robots and
  Systems (IROS)}, pages 1935--1942, 2015.

\bibitem{kitti1}
Andreas Geiger, Philip Lenz, and Raquel Urtasun.
\newblock Are we ready for autonomous driving? the kitti vision benchmark
  suite.
\newblock In {\em IEEE Conference on Computer Vision and Pattern Recognition
  (CVPR)}, pages 3354--3361, 2012.

\bibitem{slam2}
Ruben Gomez-Ojeda, Francisco-Angel Moreno, David Zu{\~n}iga-No{\"e}l, Davide
  Scaramuzza, and Javier Gonzalez-Jimenez.
\newblock Pl-slam: A stereo slam system through the combination of points and
  line segments.
\newblock {\em IEEE Transactions on Robotics}, 35(3):734--746, 2019.

\bibitem{cascade}
Xiaodong Gu, Zhiwen Fan, Siyu Zhu, Zuozhuo Dai, Feitong Tan, and Ping Tan.
\newblock Cascade cost volume for high-resolution multi-view stereo and stereo
  matching.
\newblock In {\em IEEE Conference on Computer Vision and Pattern Recognition
  (CVPR)}, pages 2495--2504, 2020.

\bibitem{gwcnet}
Xiaoyang Guo, Kai Yang, Wukui Yang, Xiaogang Wang, and Hongsheng Li.
\newblock Group-wise correlation stereo network.
\newblock In {\em IEEE Conference on Computer Vision and Pattern Recognition
  (CVPR)}, pages 3273--3282, 2019.

\bibitem{Semi-supervised}
Rongrong Ji, Ke Li, Yan Wang, Xiaoshuai Sun, Feng Guo, Xiaowei Guo, Yongjian
  Wu, Feiyue Huang, and Jiebo Luo.
\newblock Semi-supervised adversarial monocular depth estimation.
\newblock {\em IEEE transactions on pattern analysis and machine
  intelligence(TPAMI)}, 42(10):2410--2422, 2019.

\bibitem{gcnet}
Alex Kendall, Hayk Martirosyan, Saumitro Dasgupta, Peter Henry, Ryan Kennedy,
  Abraham Bachrach, and Adam Bry.
\newblock End-to-end learning of geometry and context for deep stereo
  regression.
\newblock In {\em IEEE International Conference on Computer Vision (ICCV)},
  pages 66--75, 2017.

\bibitem{iresnet}
Zhengfa Liang, Yiliu Feng, Yulan Guo, Hengzhu Liu, Wei Chen, Linbo Qiao, Li
  Zhou, and Jianfeng Zhang.
\newblock Learning for disparity estimation through feature constancy.
\newblock In {\em IEEE Conference on Computer Vision and Pattern Recognition
  (CVPR)}, pages 2811--2820, 2018.

\bibitem{mcvmfc}
Zhengfa Liang, Yulan Guo, Yiliu Feng, Wei Chen, Linbo Qiao, Li Zhou, Jianfeng
  Zhang, and Hengzhu Liu.
\newblock Stereo matching using multi-level cost volume and multi-scale feature
  constancy.
\newblock In {\em IEEE transactions on pattern analysis and machine
  intelligence (TPAMI)}, 2019.

\bibitem{receptivefield}
Wenjie Luo, Yujia Li, Raquel Urtasun, and Richard Zemel.
\newblock Understanding the effective receptive field in deep convolutional
  neural networks.
\newblock In {\em Advances in neural information processing systems (NIPS)},
  pages 4898--4906, 2016.

\bibitem{dispnet}
Nikolaus Mayer, Eddy Ilg, Philip Hausser, Philipp Fischer, Daniel Cremers,
  Alexey Dosovitskiy, and Thomas Brox.
\newblock A large dataset to train convolutional networks for disparity,
  optical flow, and scene flow estimation.
\newblock In {\em IEEE Conference on Computer Vision and Pattern Recognition
  (CVPR)}, pages 4040--4048, 2016.

\bibitem{kitti2}
Moritz Menze and Andreas Geiger.
\newblock Object scene flow for autonomous vehicles.
\newblock In {\em IEEE Conference on Computer Vision and Pattern Recognition
  (CVPR)}, pages 3061--3070, 2015.

\bibitem{nlcanet}
Zhibo Rao, Mingyi He, Yuchao Dai, Zhidong Zhu, Bo Li, and Renjie He.
\newblock Nlca-net: a non-local context attention network for stereo matching.
\newblock {\em APSIPA Transactions on Signal and Information Processing
  (ATSIP)}, 9:1--13, 2020.

\bibitem{unet}
Olaf Ronneberger, Philipp Fischer, and Thomas Brox.
\newblock U-net: Convolutional networks for biomedical image segmentation.
\newblock In {\em International Conference on Medical image computing and
  computer-assisted intervention}, pages 234--241, 2015.

\bibitem{mid}
Daniel Scharstein, Heiko Hirschm{\"u}ller, York Kitajima, Greg Krathwohl, Nera
  Ne{\v{s}}i{\'c}, Xi Wang, and Porter Westling.
\newblock High-resolution stereo datasets with subpixel-accurate ground truth.
\newblock In {\em German conference on pattern recognition (GCPR)}, pages
  31--42, 2014.

\bibitem{scharstein2002taxonomy}
Daniel Scharstein and Richard Szeliski.
\newblock A taxonomy and evaluation of dense two-frame stereo correspondence
  algorithms.
\newblock {\em International journal of computer vision (IJCV)}, 47(1-3):7--42,
  2002.

\bibitem{eth3d}
Thomas Schops, Johannes~L Schonberger, Silvano Galliani, Torsten Sattler,
  Konrad Schindler, Marc Pollefeys, and Andreas Geiger.
\newblock A multi-view stereo benchmark with high-resolution images and
  multi-camera videos.
\newblock In {\em IEEE Conference on Computer Vision and Pattern Recognition
  (CVPR)}, pages 3260--3269, 2017.

\bibitem{rank}
Markus Schulze.
\newblock A new monotonic, clone-independent, reversal symmetric, and
  condorcet-consistent single-winner election method.
\newblock {\em Social Choice and Welfare}, 36(2):267--303, 2011.

\bibitem{msmdnet}
Zhelun Shen, Yuchao Dai, and Zhibo Rao.
\newblock Msmd-net: Deep stereo matching with multi-scale and multi-dimension
  cost volume.
\newblock {\em arXiv preprint arXiv:2006.12797}, 2020.

\bibitem{pwcnet}
Deqing Sun, Xiaodong Yang, Ming-Yu Liu, and Jan Kautz.
\newblock Pwc-net: Cnns for optical flow using pyramid, warping, and cost
  volume.
\newblock In {\em IEEE Conference on Computer Vision and Pattern Recognition
  (CVPR)}, pages 8934--8943, 2018.

\bibitem{hitnet}
Vladimir Tankovich, Christian H{\"a}ne, Sean Fanello, Yinda Zhang, Shahram
  Izadi, and Sofien Bouaziz.
\newblock Hitnet: Hierarchical iterative tile refinement network for real-time
  stereo matching.
\newblock {\em arXiv preprint arXiv:2007.12140}, 2020.

\bibitem{madnet}
Alessio Tonioni, Fabio Tosi, Matteo Poggi, Stefano Mattoccia, and Luigi~Di
  Stefano.
\newblock Real-time self-adaptive deep stereo.
\newblock In {\em IEEE Conference on Computer Vision and Pattern Recognition
  (CVPR)}, pages 195--204, 2019.

\bibitem{sungeneralizaiton}
Jialiang Wang, Varun Jampani, Deqing Sun, Charles Loop, Stan Birchfield, and
  Jan Kautz.
\newblock Improving deep stereo network generalization with geometric priors.
\newblock {\em arXiv preprint arXiv:2008.11098}, 2020.

\bibitem{self-supervised}
Jamie Watson, Oisin~Mac Aodha, Daniyar Turmukhambetov, Gabriel~J. Brostow, and
  Michael Firman.
\newblock Learning stereo from single images.
\newblock In {\em European Conference on Computer Vision ({ECCV})}, pages
  722--740, 2020.

\bibitem{aanet}
Haofei Xu and Juyong Zhang.
\newblock Aanet: Adaptive aggregation network for efficient stereo matching.
\newblock In {\em IEEE Conference on Computer Vision and Pattern Recognition
  (CVPR)}, pages 1959--1968, 2020.

\bibitem{hsm}
Gengshan Yang, Joshua Manela, Michael Happold, and Deva Ramanan.
\newblock Hierarchical deep stereo matching on high-resolution images.
\newblock In {\em IEEE Conference on Computer Vision and Pattern Recognition
  (CVPR)}, pages 5515--5524, 2019.

\bibitem{cvpmvsnet}
Jiayu Yang, Wei Mao, Jose~M Alvarez, and Miaomiao Liu.
\newblock Cost volume pyramid based depth inference for multi-view stereo.
\newblock In {\em IEEE Conference on Computer Vision and Pattern Recognition
  (CVPR)}, pages 4877--4886, 2020.

\bibitem{hd3}
Zhichao Yin, Trevor Darrell, and Fisher Yu.
\newblock Hierarchical discrete distribution decomposition for match density
  estimation.
\newblock In {\em IEEE Conference on Computer Vision and Pattern Recognition
  (CVPR)}, pages 6044--6053, 2019.

\bibitem{ganet}
Feihu Zhang, Victor Prisacariu, Ruigang Yang, and Philip~HS Torr.
\newblock Ga-net: Guided aggregation net for end-to-end stereo matching.
\newblock In {\em IEEE Conference on Computer Vision and Pattern Recognition
  (CVPR)}, pages 185--194, 2019.

\bibitem{dsmnet}
Feihu Zhang, Xiaojuan Qi, Ruigang Yang, Victor Prisacariu, Benjamin Wah, and
  Philip Torr.
\newblock Domain-invariant stereo matching networks.
\newblock In {\em Europe Conference on Computer Vision (ECCV)}, pages 420--439,
  2020.

\bibitem{acfnet}
Youmin Zhang, Yimin Chen, Xiao Bai, Suihanjin Yu, Kun Yu, Zhiwei Li, and
  Kuiyuan Yang.
\newblock Adaptive unimodal cost volume filtering for deep stereo matching.
\newblock In {\em Proceedings of the AAAI Conference on Artificial Intelligence
  (AAAI)}, pages 12926--12934, 2020.

\bibitem{self-supervised-stereo}
Yiran Zhong, Yuchao Dai, and Hongdong Li.
\newblock Self-supervised learning for stereo matching with self-improving
  ability.
\newblock {\em CoRR}, abs/1709.00930, 2017.

\end{thebibliography}
}

\end{document}